\documentclass{article}
\usepackage{arxiv}
\usepackage[utf8]{inputenc} 
\usepackage[T1]{fontenc}    
\usepackage{hyperref}       
\usepackage{url}            
\usepackage{booktabs}       
\usepackage{amsfonts}       
\usepackage{nicefrac}       
\usepackage{microtype}      
\usepackage{graphicx}
\usepackage{natbib}
\usepackage{doi}

\usepackage{microtype}
\usepackage{graphicx}
\usepackage{subfigure}
\usepackage{booktabs} 

\usepackage{hyperref}

\usepackage{amsmath}
\usepackage{amssymb}
\usepackage{mathtools}
\usepackage{amsthm}

\usepackage[capitalize,noabbrev]{cleveref}

\usepackage{amsmath,amsfonts,bm}

\def\eqref#1{equation~\ref{#1}}

\def\1{\bm{1}}

\def\rve{{\mathbf{e}}}

\def\rvv{{\mathbf{v}}}

\def\rvx{{\mathbf{x}}}

\def\rmD{{\mathbf{D}}}
\def\rmE{{\mathbf{E}}}

\def\rmV{{\mathbf{V}}}

\def\rmX{{\mathbf{X}}}

\def\vy{{\bm{y}}}

\DeclareMathAlphabet{\mathsfit}{\encodingdefault}{\sfdefault}{m}{sl}
\SetMathAlphabet{\mathsfit}{bold}{\encodingdefault}{\sfdefault}{bx}{n}

\def\gG{{\mathcal{G}}}

\def\gP{{\mathcal{P}}}

\def\sX{{\mathbb{X}}}

\newcommand{\R}{\mathbb{R}}

\newcommand{\parents}{Pa} 

\usepackage{algorithm}
\usepackage[normalem]{ulem}
\useunder{\uline}{\ul}{}
\usepackage{wrapfig}
\usepackage{booktabs}
\usepackage{multirow}
\usepackage{graphicx}
\usepackage{caption}
\usepackage{array}
\usepackage{longtable}
\newcolumntype{C}[1]{>{\centering\arraybackslash}p{#1}}
\usepackage{afterpage}
\usepackage{algorithmic}

\theoremstyle{plain}

\theoremstyle{definition}

\theoremstyle{remark}

\newcommand{\eat}[1]{}

\newcommand{\yu}[1]{{\color{red} [Yu: #1] }}
\newcommand{\sva}[1]{{\color{orange} [Shivvrat: #1] }}
\newcommand{\icml}[1]{{\color{blue} [{\color{purple}\textbf{Changes :}} #1]}}

\newcommand{\vibhav}[1]{{\color{orange} [Vibhav: #1] }}

\renewcommand{\sva}[1]{}
\renewcommand{\icml}[1]{#1}
\renewcommand{\vibhav}[1]{}
\renewcommand{\yu}[1]{}

\newcommand{\cnn}[1]{CNN}
\newcommand{\mrfgs}[1]{DRF - GS}
\newcommand{\ilp}[1]{DRF - ILP}
\newcommand{\ijgp}[1]{DRF - IJGP}
\newcommand{\dnlr}[1]{DDN - LR - Pipeline}
\newcommand{\dnlrjoint}[1]{DDN - LR - Joint}
\newcommand{\dnnn}[1]{DDN - MLP - Pipeline}
\newcommand{\dnnnjoint}[1]{DDN - MLP - Joint}

\usepackage[textsize=tiny]{todonotes}

\title{Deep Dependency Networks for Multi-Label Classification}

\date{}

\author{Shivvrat Arya \\
	Department of Computer Science\\
	The University of Texas at Dallas\\
	Dallas, TX 75252 \\
	\texttt{shivvrat.arya@utdallas.edu} \\
	\And
	Yu Xiang \\
	Department of Computer Science\\
	The University of Texas at Dallas\\
	Dallas, TX 75252 \\
	\texttt{yu.xiang@utdallas.edu}  \\
 	\And
	Vibhav Gogate \\
	Department of Computer Science\\
	The University of Texas at Dallas\\
	Dallas, TX 75252 \\
	\texttt{vibhav.gogate@utdallas.edu}  \\
}

\hypersetup{
pdftitle={Deep Dependency Networks for Multi-Label Classification},
pdfsubject={cs.AI, cs.LG},
pdfauthor={Shivvrat Arya, Yu Xiang, Vibhav Gogate},
pdfkeywords={Multi-label Classification, Probabilistic Graphical Models, Multi-label Action Classification, Multi-label Image Classification, Dependency Networks},
}

\begin{document}
\maketitle
\begin{abstract}
We propose a simple approach which combines the strengths of probabilistic graphical models and deep learning architectures for solving the multi-label classification task, focusing specifically on image and video data.  First, we show that the performance of previous approaches that combine Markov Random Fields with neural networks can be modestly improved by leveraging more powerful methods such as iterative join graph propagation, integer linear programming, and $\ell_1$ regularization-based structure learning. Then we propose a new modeling framework called \textit{deep dependency networks}, which augments a dependency network, a model that is easy to train and learns more accurate dependencies but is limited to Gibbs sampling for inference, to the output layer of a neural network. We show that despite its simplicity, jointly learning this new architecture yields significant improvements in performance over the baseline neural network. In particular, our experimental evaluation on three video activity classification datasets: Charades, Textually Annotated Cooking Scenes (TACoS), and Wetlab, and three multi-label image classification datasets: MS-COCO, PASCAL VOC, and NUS-WIDE show that deep dependency networks are almost always superior to pure neural architectures that do not use dependency networks.

\end{abstract}
\section{Introduction}
\label{sec:intro}

In this paper, we focus on the multi-label classification (MLC) task, and more specifically on its two notable instantiations, multi-label action classification (MLAC) for videos and multi-label image classification (MLIC). At a high level, given a pre-defined set of labels (or actions) and a test example (video or image), the goal is to assign each test example to a subset of labels.
It is well known that MLC is notoriously difficult because in practice the labels are often correlated, and thus predicting them independently may lead to significant errors. Therefore, most advanced methods explicitly model the relationship or dependencies between the labels, using either probabilistic techniques \citep{wang2008generative,rossiProceedingsTwentyThirdInternational2013b,antonucci2013ensemble,wangEnhancingMultilabelClassification2014a,Tan_2015_CVPR,pmlr-v52-dimauro16} or non-probabilistic/neural methods  \citep{kongMultilabelClassificationMining2013a,papagiannopoulouDiscoveringExploitingDeterministic2015a,chenLearningSemanticSpecificGraph2019,chenMultiLabelImageRecognition2019,wang2021semi,nguyenModularGraphTransformer2021,wangNovelReasoningMechanism2021a,liuQuery2LabelSimpleTransformer2021, quMultilayeredSemanticRepresentation2021}.

To this end, motivated by approaches that combine probabilistic graphical models (PGMs) with neural networks (NNs)  \citep{krishnanDeepKalmanFilters2015, johnsonComposingGraphicalModels2017}, as a starting point, we investigated using (Conditional) Markov random fields (CRFs and MRFs), a type of undirected PGM, to capture the relationship between the labels as well as those between the labels and features derived from feature extractors. Unlike previous work, which used these MRF+NN or CRF+NN hybrids with conventional inference schemes such as Gibbs sampling (GS) and mean-field inference, our goal was to evaluate whether \textit{advanced approaches}, specifically (1) iterative join graph propagation (IJGP) \citep{mateescu2010join}, a type of generalization Belief propagation technique \citep{yedidiaGeneralizedBeliefPropagation2000}, (2) integer linear programming (ILP) based techniques for computing most probable explanations and (3) a well-known structure learning method based on logistic regression with $\ell_1$-regularization \citep{NIPS2006_lee_a4380923,NIPS2006_Wainwright_86b20716}, can improve the generalization performance of MRF+NN hybrids. 

To measure and compare performance of these MRF+NN hybrids with NN models, we used several metrics such as mean average precision (mAP), label ranking average precision (LRAP), subset accuracy (SA), and the jaccard index (JI) and experimented on three video datasets: (1) Charades \citep{charades_paper}, (2) TACoS \citep{tacos:regnerietal:tacl} and (3) Wetlab \citep{wetlab:Naim2014UnsupervisedAO} and three image datasets: (1) MS-COCO \cite{linMicrosoftCOCOCommon2015}, (2) PASCAL VOC 2007 \cite{everinghamPascalVisualObject2010} and (3) NUS-WIDE \cite{chuaNUSWIDERealworldWeb2009a}. We found that generally speaking, both IJGP and ILP are superior to the baseline NN and Gibbs sampling in terms of JI and SA but are sometimes inferior to the NN in terms of mAP and LRAP. We speculated that because MRF structure learners only allow pairwise relationships and impose sparsity or low-treewidth constraints for faster, accurate inference, they often yield poor posterior probability estimates in high-dimensional settings. Since both mAP and LRAP require good posterior probability estimates, GS, IJGP, and ILP exhibit poor performance when mAP and LRAP are used to evaluate the performance.

To circumvent this issue and in particular to derive good posterior estimates, we propose a new PGM+NN hybrid called \textit{deep dependency networks} (DDNs). At a high level, a dependency network (DN) \citep{heckerman2000dependency} represents a joint probability distribution using a collection of conditional distributions, each defined over a variable (label) given all other variables (labels) in the network. Because each conditional distribution can be trained locally, DNs are easy to train. However, a caveat is that they are limited to Gibbs sampling for inference and are not amenable to advanced probabilistic inference techniques \citep{lowdClosedFormLearningMarkov2012a}. 

\begin{figure}[t]
\vskip 0.2in
        \begin{center}
        \centerline{\includegraphics[width=0.9\columnwidth]{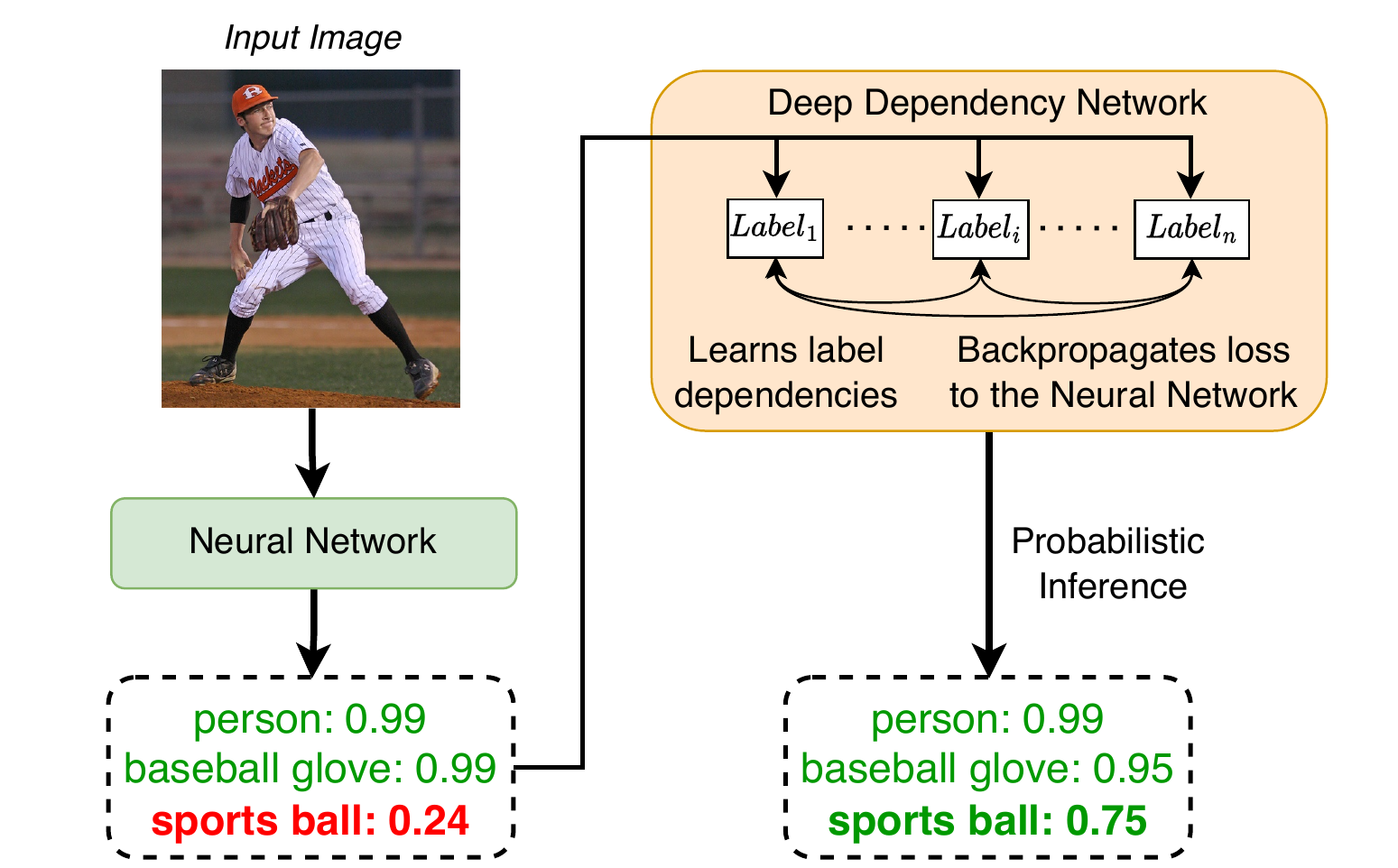}}
        \caption{\icml{Illustration of improvements made by our proposed deep dependency network for  multi-label image classification. The DDN learns label relationships, backpropagates the loss that reasons about label relationships to the neural network (NN) and helps to predict labels missed by the NN, such as the \textbf{sports ball} (occluded object).}}
        \label{fig:ddn_intro}
        \end{center}
\vspace{-8mm}
\end{figure}
In our proposed deep dependency network (DDN) architecture, a dependency network sits on top of a feature extractor based on a neural network. \icml{We illustrate the workings of the DDN architecture in Figure \ref{fig:ddn_intro}}. The feature extractor converts the input image or video segment to a set of features, and the dependency network uses these features to define a local conditional distribution over each label given the features and other labels. We show that deep dependency models are easy to train either jointly or via a pipeline method where the feature extractor is trained first, followed by the DNN by defining an appropriate loss function that minimizes the negative pseudo log-likelihood \citep{besag_statistical_1975} of the data. 
We conjecture that because DDNs can be quite dense, they often learn a better representation of the data, and as a result, they are likely to outperform MRFs learned from data in terms of posterior predictions. 

We trained DDNs using the pipeline and joint learning approaches and evaluated them using the four aforementioned metrics (JI, SA, mAP and LRAP) and six datasets. We observed that jointly trained DDNs are often superior to the baseline neural networks as well as advanced MRF+NN methods that use GS, IJGP, and ILP on all four metrics. Specifically, they achieve the highest scores on all metrics on five out of the six datasets. Also, we found that the jointly trained DDNs are more accurate than the ones trained using the pipeline approach. This is because end-to-end (or joint) learning steers the feature selection process to provide tailored features for the DN, and learns a DN that is in-turn tailored to the output of this backbone. DDNs provide a multi-label classification head that works on the features extracted by the backbone.

In summary, this paper makes the following contributions:
\begin{itemize}
    \item We propose a new hybrid model called deep dependency networks that combines the strengths of dependency networks (faster training and access to probabilistic inference schemes) and neural networks (flexibility, high-quality feature representation).
    \item We experimentally evaluate DDNs on three video datasets and three image datasets by using four metrics for solving the multi-label action classification and multi-label image classification tasks. This helps us to show that DDNs can be used for diverse multi-label classification tasks. We found that jointly trained DDNs consistently outperform NNs and MRF+NN hybrids on all metrics and datasets. 
\end{itemize}

\section{Preliminaries}
\label{sec:prelim}
A \textbf{log-linear model} or a \textbf{Markov random field} (MRF), denoted by $\mathcal{M}$, is an undirected probabilistic graphical model \citep{koller2009probabilistic} that is widely used in many real-world domains for representing and reasoning about uncertainty. It is defined as a triple $\langle \rmX, \mathcal{F}, \Theta \rangle$ where $\rmX{}=\{X_1,\ldots,X_n\}$ is a set of Boolean random variables, $\mathcal{F}=\{f_1,\ldots,f_m\}$ is a set of features such that each feature $f_i$ (we assume that a feature is a Boolean formula) is defined over a subset $\rmD_i$ of $\rmX$, and $\Theta=\{\theta_1,\ldots,\theta_m\}$ is a set of real-valued weights or parameters, namely $\forall \theta_i \in \Theta;\; \theta_i \in \R$ such that each feature $f_i$ is associated with a parameter $\theta_i$. $\mathcal{M}$ represents the following probability distribution:
\begin{equation}
    P(\rvx)=\frac{1}{Z(\Theta)} \exp \left\{\sum_{i=1}^m \theta_i f_i\left(\rvx_{\rmD_i}\right)\right\}
\label{eq:log_linear_pd}
\end{equation}
where $\rvx$ is an assignment of values to all variables in $\rmX$, $\rvx_{\rmD_i}$ is the projection of $\rvx$ on the variables $\rmD_i$ of $f_i$, $f_i(\rvx_{\rmD_i})$ is an \textit{indicator function} that equals $1$ when the assignment $\rvx_{\rmD_i}$ evaluates $f_i$ to \texttt{True} and is $0$ otherwise, and $Z(\Theta)$ is the normalization constant called the \textit{partition function}.

We focus on three tasks over MRFs: (1) structure learning which is the problem of learning the features and parameters given training data; (2) posterior marginal inference which is the task of computing the marginal probability distribution over each variable in the network given evidence (evidence is defined as an assignment of values to a subset of variables); and (3) finding the most likely assignment to all the non-evidence variables given evidence (this task is often called maximum-a-posteriori or MAP inference in short). All of these tasks are at least NP-hard in general and therefore approximate methods are often preferred over exact ones in practice.

A popular and fast method for structure learning is to learn binary pairwise MRFs by training an $\ell_1$-regularized logistic regression classifier for each variable given all other variables as features \citep{NIPS2006_Wainwright_86b20716,NIPS2006_lee_a4380923}. $\ell_1$-regularization induces sparsity in that it encourages many weights to take the value zero. All non-zero weights are then converted into conjunctive features. Each conjunctive feature evaluates to $\texttt{True}$ if both variables are assigned the value $1$ and to $\texttt{False}$ otherwise. Popular approaches for posterior marginal inference are the Gibbs sampling algorithm and generalized Belief propagation \citep{yedidiaGeneralizedBeliefPropagation2000} techniques such as Iterative Join Graph Propagation \citep{mateescu2010join}. For MAP inference, a popular approach is to encode the optimization problem as a linear integer programming problem  \citep{koller2009probabilistic} and then use off-the-shelf approaches such as \citep{gurobi} to solve the latter.

\textbf{Dependency Networks (DNs)} \citep{heckerman2000dependency} represent the joint distribution using a set of local conditional probability distributions, one for each variable. Each conditional distribution defines the probability of a variable given all of the others. A DN is consistent if there exists a joint probability distribution $P(\rvx)$ such that all conditional distributions $P_i(x_i|\rvx_{-i})$ where $\rvx_{-i}$ is the projection of $\rvx$ on $\rmX \setminus 
\{X_i\}$, are conditional distributions of $P(\rvx)$. 

A DN is learned from data by learning a  classifier (e.g., logistic regression, multi-layer perceptron, etc.) for each variable, and thus DN learning is embarrassingly parallel. However, because the classifiers are independently learned from data, we often get an inconsistent DN. It has been conjectured \citep{heckerman2000dependency} that most DNs learned from data are almost consistent in that only a few parameters need to be changed in order to make them consistent.

The most popular inference method over DNs is \textit{fixed-order} Gibbs sampling \citep{liubook}. If the DN is consistent, then its conditional distributions are derived from a
joint distribution $P(\rvx)$, and the stationary distribution (namely the distribution that Gibbs sampling converges to) will be the same as $P(\rvx)$. If the DN is inconsistent, then the
stationary distribution of Gibbs sampling will be inconsistent with the conditional distributions.

\section{Deep Dependency Networks}
\label{sec:ddn}
\icml{In this section, we describe how to solve the multi-label action classification task in videos and the multi-label image classification task using a hybrid of dependency networks and neural networks. At a high level, the neural network provides high-quality features given video segments/images and the dependency network represents and reasons about the relationships between the labels and features.}
\eat{In this section, we describe how to solve the multi-label action classification task in videos using a hybrid of dependency networks and neural networks. At a high level, the neural network provides high-quality features given videos and the dependency network represents and reasons about the probabilistic relationships between the labels and features.}

\subsection{Framework}
\begin{figure}[t]
\begin{center}
\centerline{\includegraphics[width=.75\textwidth]{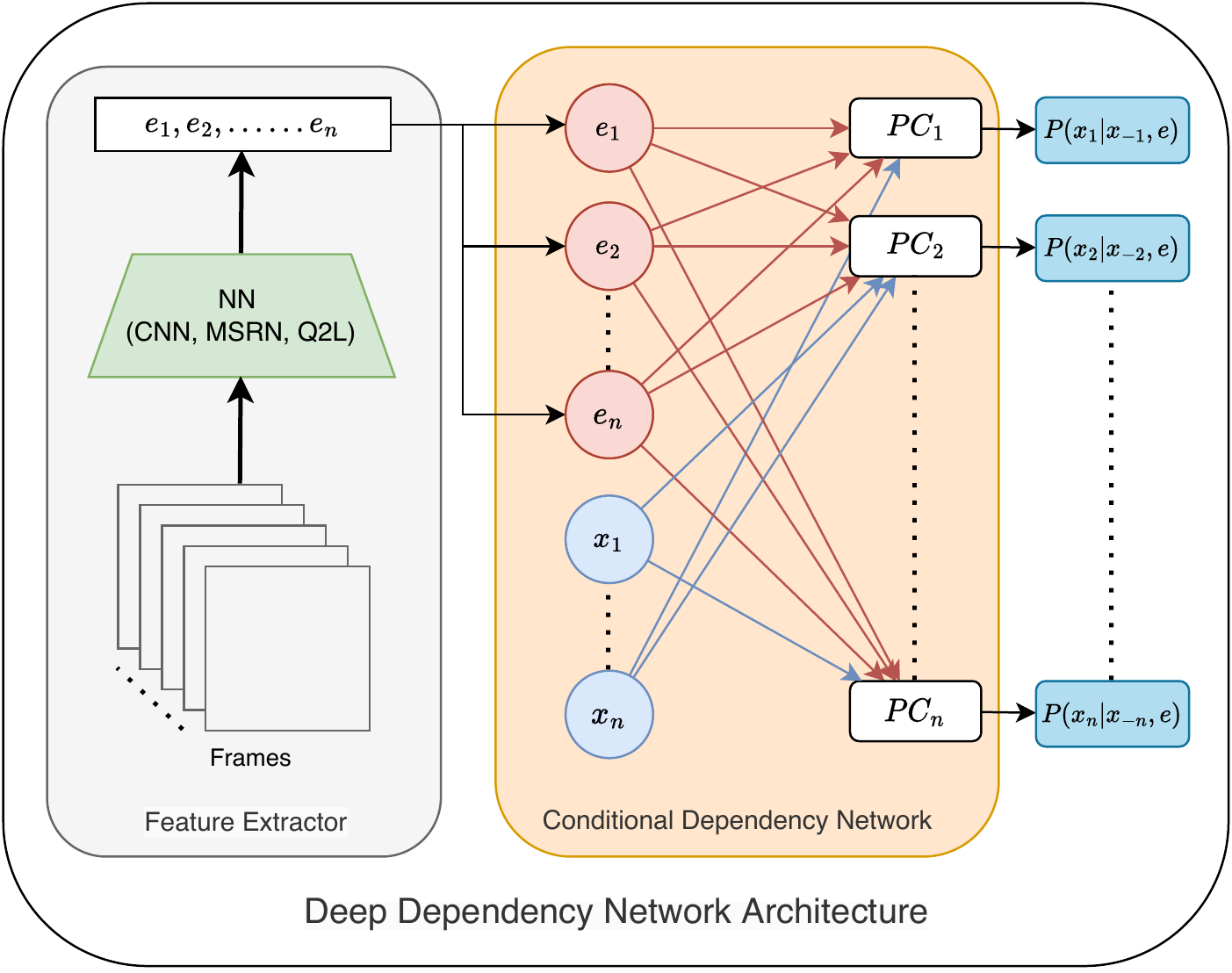}}
        \caption{\icml{Illustration of Dependency Network for multi-label video classification. The NN takes video clips (frames) as input and outputs the features $E_1,E_2,...,E_n$ (denoted by red colored nodes) for the DN. These features are then used by the probabilistic classifiers ($PC_1$, $\ldots$, $PC_n$) to model the local conditional distributions. At each output node (blue boxes), the form of the conditional distribution is variable given its parents (incoming arrows represented by orange and blue color) in the conditional dependency network.}}
        \label{fig:ddn}
        \end{center}
\vspace{-8mm}
\end{figure}
Let $\rmV$ denote the set of random variables corresponding to the pixels and $\rvv$ denote the RGB values of the pixels in a frame or a video segment. Let $\rmE$ denote the (continuous) output nodes of a neural network which represents a function $\mathbb{N}: \rvv \rightarrow \rve$, that takes $\rvv$ as input and outputs an assignment $\rve$ to $\rmE$. Let $\rmX=\{X_1,\ldots,X_n\}$ denote the set of labels (actions). For simplicity, we assume that $|\rmE|=|\rmX|=n$. Given $(\rmV,\rmE,\rmX)$, a deep dependency network (DDN) is a pair $ \langle \mathcal{N},\mathcal{D} \rangle$ where $\mathcal{N}$ is a neural network that maps $\rmV=\rvv$ to $\rmE=\rve$ and $\mathcal{D}$ is a conditional dependency network \citep{guo2011multi} that models 
$P(\rvx|\rve)$ where $\rve=\mathbb{N}(\rvv)$. The conditional dependency network represents the distribution $P(\rvx|\rve)$ using a collection of local conditional distributions $P_i(x_i|\rvx_{-i},\rve)$, one for each \icml{label}\eat{action (label)} $X_i$, where $\rvx_{-i} = \{ x_1, \ldots, x_{i-1}, x_{i+1}, \ldots, x_{n}\}$. 

Thus, a DDN is a discriminative model and represents the conditional distribution  $P(\rvx|\rvv)$ using several local conditional distributions $P(x_i|\rvx_{-i},\rve)$ and makes the following conditional independence assumptions $P(x_i|\rvx_{-i},\rvv)=P(x_i|\rvx_{-i},\rve)$ where $\rve=\mathbb{N}(\rvv)$. \icml{Figure~ \ref{fig:ddn} shows the architecture of a DDN for solving the multi-label action classification task in videos.}


\subsection{Learning}
We can either train the DDN using a \textit{pipeline method} or via \textit{joint training}. In the pipeline method, we first train the neural network using standard approaches (e.g., using cross-entropy loss) or  use a pre-trained model. Then for each training example $(\rvv,\rvx)$, we send the video/image through the neural network to obtain a new representation $\rve$ of $\rvv$. The aforementioned process transforms each training example $(\rvv,\rvx)$ into a new feature representation $(\rve,\rvx)$ where $\rve=\mathbb{N}(\rvv)$. Finally, for each label $X_i$, we learn a classifier to model the conditional distribution $P_i(x_i|\rvx_{-i},\rve)$. Specifically, given a training example $(\rve,\rvx)$, each probabilistic classifier indexed by $i$ \icml{($PC_i$)}, uses $X_i$ as the class variable and $(\rmE \cup \rmX_{-i})$ as the attributes. In our experiments, we used two probabilistic classifiers, logistic regression and multi-layer perceptron.

The pipeline method has several useful properties: it requires modest computational resources, is relatively fast and can be easily parallelized. As a result, it is especially beneficial when (only) less powerful GPUs are available at training time but a pre-trained network that is trained using more powerful GPUs is readily available. 

\eat{\vibhav{Need to change the stuff below. It does not sound right. The loss as presented makes no intuitive sense. 
To perform \textit{joint training}, we propose to use negative conditional pseudo log-likelihood \cite{besag_statistical_1975} as our loss function.

We can say something like; easiest loss to use is cross entropy at the dependency net layer. However, its performance was not good. The issue is that it will get rid of the original intended meaning of the output nodes in the neural network; the goal of each node was also to predict the action. Therefore, just at the output node, we also add the original output loss. 
}}
For joint learning, we propose to use the conditional pseudo log-likelihood loss (CPLL)  \cite{besag_statistical_1975}. Let $\Theta$ denote the set of parameters of the DDN, then the CPLL is given by
\begin{align}
\mathcal{L}(\Theta,\rvv,\rvx)&= - \sum_{i=1}^{n} \log P_i(x_i|\rvv,\rvx_{-i}; \Theta) \\
\label{eq:loss1}
&=- \sum_{i=1}^{n}\log P_i(x_i|\rve=\mathbb{N}(\rvv),\rvx_{-i}; \Theta)
\end{align}
In practice, for faster training/convergence, we will partition the parameters $\Theta$ of the DDN into two (disjoint) subsets $\Pi$ and $\Gamma$ where $\Pi$  and $\Gamma$ denote the parameters of the neural network and local conditional distributions respectively; and initialize $\Pi$ using a pre-trained neural network and $\Gamma$ using the pipeline method. Then, we can use any gradient-based (backpropagation) method to minimize the loss function.

\subsection{Inference: Using the DDN to Make Predictions}
Unlike a conventional discriminative model such as a neural network, in a DDN, we cannot predict the output labels by simply making a forward pass over the network. This is because each probabilistic classifier indexed by $i$ (which yields a probability distribution over $X_i$) requires an assignment $\rvx_{-i}$ to all labels except $x_i$, and $\rvx_{-i}$ is not available at prediction time.   

To address this issue, we use the following Gibbs sampling based approach (a detailed algorithm is provided in the appendix). 
We first send the \icml{frame/segment}  $\rvv$ through the neural network $\mathcal{N}$ to yield an assignment $\rve$ to the output nodes of the neural network. Then, we perform fixed-order Gibbs Sampling over the dependency network where the latter represents the distribution $P(\rvx|\rve)$. Finally, given samples $(\rvx^{(1)},\ldots,\rvx^{(N)})$ generated via Gibbs sampling, we estimate the marginal probability distribution of each label $X_i$ using the following mixture estimator \citep{liubook}:
\begin{align}
\label{eq:Mixture_Estimator1}
\hat{P}_i\left(x_{i}|\rvv\right)=\frac{1}{N} \sum_{j=1}^{N} P_i\left(x_{i} \mid \rvx_{-i}^{(j)},\rve\right)
\end{align}

\sva{Should we add a few more things to this section? It looks smaller for the methods section of the paper. Otherwise we can also change the name of the previous section from preliminaries to something else!}

\section{Experimental Evaluation}
\label{sec:experiments}
\icml{In this section, we evaluate the proposed models on two multi-label classification tasks: (1) multi-label activity classification using three video datasets; and (2) multi-label image classification using three image datasets.} We begin by describing the datasets and metrics, followed by the experimental setup, and conclude with the results. All models were implemented using PyTorch, and one NVIDIA A40 GPU was used to train and test all the models. 
\subsection{Datasets and Metrics}
\label{sec:datasets}
We evaluated our algorithms on the following three video datasets: (1) Charades \citep{charades_paper}; (2) Textually Annotated Cooking Scenes (TACoS) \citep{tacos:regnerietal:tacl}; and (3) Wetlab \citep{naim-etal-2015-discriminative}. In the Charades dataset, the videos are divided into segments (video clips), and each segment is annotated with one or more action labels. In the TACoS and Wetlab datasets, each frame is associated with one or more actions. 

\textbf{Charades dataset} \citep{charades_paper} comprises of videos of people performing daily indoor activities while interacting with various objects.\eat{It is a multi-label activity classification dataset; thus, more than one activity can happen at a given time.} In the standard split, there are 7,986 training videos and 1,863 validation videos\eat{, averaging 30 seconds}. We used the training videos to train the models and the validation videos for testing purposes. We follow the instructions provided in PySlowFast \citep{fan2020pyslowfast} to do the train-test split for the dataset. The dataset has roughly 66,500 temporal annotations for 157 action classes. \eat{We report mean Average Precision (mAP) following the standard evaluation protocols for this dataset. We also report Label Ranking Average Precision (LRAP) and Jaccard Index (JI), as these metrics have been used extensively in previous work for evaluating the performance of multi-label classifiers.} 



The \textbf{\eat{The Textually Annotated Cooking Scenes (}TaCOS dataset} \citep{tacos:regnerietal:tacl} consists of third-person videos of a person cooking in a kitchen. The dataset comes with hand-annotated labels of actions, objects, and locations for each video frame. From the complete set of these labels, we selected 28 labels. 
\eat{Each of these labels corresponds to a location, an object, or a verb, and each action is defined as a (location, object, verb) triple. By selecting the videos that correspond to these labels and}
By dividing the videos corresponding to these labels into train and test sets, we get a total of 60,313 frames for training and 9,355 frames for testing, spread out over 17 videos. 

The \textbf{Wetlab dataset} \citep{wetlab:Naim2014UnsupervisedAO} comprises of videos where experiments are being performed by lab technicians that involve hazardous chemicals in a wet laboratory. 
We used five videos for training and one video for testing. The training set comprises 100,054 frames, and the test set comprises 11,743 frames.\eat{Each activity in the dataset can have multiple objects per activity (for example, an activity ``transfer contents'' can have two objects linked with it, a test tube and a beaker) and there are 57 possible labels.} There are 57 possible labels and each label corresponds to an object or a verb, and each action is made of one or more labels from each category.

\icml{We also evaluated our algorithms on three multi-label image classification (MLIC) datasets: (1) MS-COCO \citep{linMicrosoftCOCOCommon2015}; (2) PASCAL VOC 2007,  and  \citep{everinghamPascalVisualObject2010};  and (3) NUS-WIDE \citep{chuaNUSWIDERealworldWeb2009a}.

\textbf{MS-COCO} (Microsoft Common Objects in Context) \citep{linMicrosoftCOCOCommon2015} is a large-scale object detection and segmentation dataset. It has also been extensively used for the MLIC task. The dataset contains 122,218 labeled images and 80 labels in total. Each image is labeled with at least 2.9 labels on average. We used the 2014 version of the dataset\eat{ as mentioned in \citep{liuQuery2LabelSimpleTransformer2021}}.

\textbf{NUS-WIDE} dataset \citep{chuaNUSWIDERealworldWeb2009a} is a real-world web image dataset that contains 269,648 images from Flickr. Each image has been manually annotated with a subset of 81 visual classes that include objects and scenes. 

\textbf{PASCAL VOC 2007} \citep{everinghamPascalVisualObject2010} is another dataset that has been used widely for the MLIC task.
The dataset contains 5,011 images in the train-validation set and 4,952 images in the test set. The total number of labels in the dataset is 20 which corresponds to object classes.\eat{ and each image can take one or more labels. }


We follow the instructions provided in \citep{quMultilayeredSemanticRepresentation2021} to do the train-test split for NUS-WIDE and PASCAL VOC 2007. 
} \icml{We evaluated the performance on the TACoS, Wetlab, MS-COCO, NUS-WIDE, and VOC datasets using the following four metrics: mean Average Precision (mAP), Label Ranking Average Precision (LRAP), Subset Accuracy (SA), and Jaccard Index (JI). For all the metrics that are being considered here, a higher value means better performance. mAP and LRAP require access to an estimate of the posterior probability distribution at each label while SA and JI are non-probabilistic and only require an assignment to the labels. 
Note that we only report mAP, LRAP and JI on the Charades dataset because existing approaches cannot achieve reasonable SA due to large size of the label space.}

In recent years, mAP has been used as an evaluation metric for multi-label image and action classification in lieu of conventional metrics such as SA and JI. However, both SA (which seeks exact match with the ground truth) and JI are critical for applications such as dialogue systems \citep{vilarMultilabelTextClassification2004}, self-driving cars \citep{mlfselfdriving, protopapadakisMultilabelDeepLearning2020} and disease diagnosis \citep{maxwellDeepLearningArchitectures2017, zhangNovelDeepNeural2019, zhouApplicationMultilabelClassification2021} where MLC is one of the sub-tasks in a series of interrelated sub-tasks. Missing a single label in these applications could have disastrous consequences for downstream sub-tasks.


\subsection{Experimental Setup and Methods}

We used three types of architectures in our experiments: (1) Baseline \icml{neural networks} which are specific to each dataset; (2) \icml{neural networks}\eat{CNNs} augmented with MRFs, which we will refer to as deep random fields or DRFs in short; and (3) the method proposed in this paper which uses a dependency network on top of the \icml{neural networks} called deep dependency networks (DDNs). 

\eat{\textbf{Neural Networks}. \icml{We choose four different types of models\eat{ for extracting the features}, and they act as a baseline for the experiments and as a feature extractor for DRFs and DDNs. We chose two types of models for this, 2D CNNs and 3D CNNs. In doing this, we want to demonstrate that the proposed method can work well for two different types of feature extractors.}}

\textbf{Neural Networks}. \icml{We choose four different types of neural networks\eat{ for the feature extraction}, and they act as a baseline for the experiments and as a feature extractor for DRFs and DDNs. Specifically, we experimented with: (1) 2D CNN, (2) 3D CNN, (3) transformers, and (4) CNN with attention module and graph attention networks (GAT) \citep{velivckovicgraph}. 
This helps us show that our proposed method can improve the performance of a wide variety of neural architectures, even those which model label relationships because unlike the latter it performs probabilistic inference (Gibbs sampling).}

 For the Charades dataset, we use the PySlowFast \citep{fan2020pyslowfast} implementation of the SlowFast Network \citep{Feichtenhofer_2019_ICCV} (a state-of-the-art 3D CNN for video classification) which uses a 3D ResNet model as the backbone. For TACoS and Wetlab datasets, we use InceptionV3 \citep{szegedyRethinkingInceptionArchitecture2016}, one of the state-of-the-art 2D CNN models for image classification. \icml{For the MS-COCO dataset, we used Query2Label (Q2L) \cite{liuQuery2LabelSimpleTransformer2021}, which uses transformers to pool class-related features. Q2L also learns label embeddings from data to capture the relationships between the labels. 
Finally, we used the  multi-layered semantic representation network (MSRN) \citep{quMultilayeredSemanticRepresentation2021} for NUS-WIDE and PASCAL VOC. MSRN also models label correlations and learns semantic representations at multiple convolutional layers. For extracting the features for Charades, MS-COCO, NUS-WIDE, and PASCAL VOC datasets, we use the pre-trained models and hyper-parameters provided in their respective repositories.} For TaCOS and Wetlab datasets, we fine-tuned an InceptionV3 model that was pre-trained on the ImageNet dataset. 

\textbf{Deep Random Fields (DRFs)}. As a baseline, we used a model that combines MRFs with \icml{neural networks}. This DRF model is similar to the DDN except that we use an MRF instead of a DDN to compute $P(\rvx|\rve)$. We trained the MRFs generatively; namely, we learned a joint distribution $P(\rvx,\rve)$, which can be used to compute $P(\rvx|\rve)$ by instantiating evidence. We chose generative learning because we learned the structure of the MRFs from data, and discriminative structure learning is slow in practice \citep{koller2009probabilistic}. Specifically, we used the logistic regression with $\ell_1$ regularization method of \citep{NIPS2006_Wainwright_86b20716} to learn a pairwise MRF. The training data for this method is obtained by sending each annotated video clip (or frame) $(\rvv,\rvx)$ through the neural network and transforming it to $(\rve,\rvx)$ where $\rve=\mathbb{N}(\rvv)$. At termination, this method yields a graph $\gG$ defined over $\rmX \cup \rmE$. 

For parameter/weight learning, we converted each edge over $\rmX \cup \rmE$ to a conjunctive feature. For example, if the method learns an edge between $X_i$ and $E_j$, we use a conjunctive feature $X_i \wedge E_j$ which is true if both $X_i$ and $E_j$ are assigned the value $1$. Then we learned the weights for each feature by maximizing the pseudo log-likelihood of the data.

For inference over MRFs, we used Gibbs sampling (GS), Iterative Join Graph Propagation (IJGP) \cite{mateescu2010join}, and Integer Linear Programming (ILP) methods. Thus, three versions of DRFs corresponding to the inference scheme were used. We refer to these schemes as DRF-GS, DRF-ILP, and DRF-IJGP, respectively. Note that IJGP and ILP are advanced schemes, and we are unaware of their use for \icml{multi-label  classification}.\eat{multi-label action classification \icml{and multi-label image classification}}
Our goal is to test whether advanced inference schemes help improve the performance of deep random fields.

\textbf{Deep Dependency Networks (DDNs)}. We experimented with four versions of DDNs: (1) DDN-LR-Pipeline; (2) DDN-MLP-Pipeline; (3) DDN-LR-Joint; and (4) DDN-MLP-Joint. The first and third versions use logistic regression (LR), while the second and fourth versions use multi-layer perceptrons (MLP) to represent the conditional distributions. The first two versions are trained using the pipeline method, while the last two versions are trained using the joint learning loss given in \eqref{eq:loss1}.

\textbf{Hyperparameters.} For DRFs, in order to learn a sparse structure (using the logistic regression with $\ell_1$ regularization method of \citep{NIPS2006_Wainwright_86b20716}), we increased the regularization constant associated with the $\ell_1$ regularization term until the number of neighbors of each node in $\gG$ is bounded between 2 and 10. We enforced this sparsity constraint in order to ensure that the inference schemes (specifically, IJGP and ILP) are accurate and the model does not overfit to the training data. IJGP, ILP, and GS are anytime methods; for each, we used a time-bound of 60 seconds per example. 

For DDNs, we used LR with $\ell_1$ regularization and MLPs with $\ell_2$ regularization. For MLP the number of hidden layers was selected from the $\{2,3,4\}$. The regularization constants for LR and MLP (chosen from the $\{0.1,0.01,0.001\}$) and the number of layers for MLP were chosen via cross-validation. For all datasets, MLP with four layers performed the best. ReLU was used for the activation function for each hidden layer and sigmoid for the outputs. For joint learning, we reduced the learning rates of both LR and MLP models by expanding on the learning rate scheduler given in PySlowFast \citep{fan2020pyslowfast}\eat{. We used the steps with relative learning rate policy, where the learning rate depended on the current epoch} and the initial learning rate was chosen in the range of $\displaystyle [10^{-3}, 10^{-5}]$.


\eat{
need to have the number of neighbors each node can have. We select values in the range $\displaystyle [2,10]$ for these hyper-parameters. If we go beyond ten, then the inference part takes much time to complete. For the structure we provide the learning rates of $\displaystyle \{0.01, 0.001\}$ and the regularization parameters with the values $\displaystyle \{0.1, 0.01, 0.001\}$. For the inference techniques, we need to provide the maximum run time for the IJGP method, which we selected from $\displaystyle \{10, 30, 60\}$ seconds. We used Neural Networks with 3 and 4 hidden layers for the distributions of dependency networks. For all the models in dependency networks, we chose the learning rates from $\displaystyle \{0.05, 0.01, 0.001\}$ and regularization parameters from $\displaystyle \{0.1, 0.01, 0.001\}$. The number of samples for inference on the dependency network were from $\displaystyle \{1000, 10000, 50000\}$. For the joint learning experimentation, we reduced the learning rates of both models by expanding on the learning rate scheduler given with PySlowFast. We use the steps with relative learning rate policy, where the learning rate depended on the current epoch and was in the range of $\displaystyle [1e^{-3}, 1e^{-5}]$. Other hyper-parameters remain the same as the pipeline models. 
}
\subsection{Results}
\begin{table*}[t]
\centering
\caption{Comparison of our methods with the baseline for MLAC task. The best/second best values are bold/underlined. The last row shows the relative improvement made by the best performing proposed method over the baseline.}
\label{tab:eval-mlac}
\resizebox{\textwidth}{!}{%
\begin{sc}
\begin{tabular}{c|ccc|cccc|cccc}
\hline
\multirow{2}{*}{Method} &
  \multicolumn{3}{c|}{Charades} &
  \multicolumn{4}{c|}{TACoS} &
  \multicolumn{4}{c}{Wetlab} \\ \cline{2-12} 
 &
  mAP ↑ &
  LRAP ↑ &
  JI ↑ &
  mAP ↑ &
  LRAP ↑ &
  SA ↑ &
  JI ↑ &
  mAP ↑ &
  LRAP ↑ &
  SA ↑ &
  JI ↑ \\ \hline
SlowFast\citep{fan2020pyslowfast} &
  0.388 &
  0.535 &
  0.294 &
   &
   &
   &
   &
   &
   &
   &
   \\
InceptionV3\citep{szegedyRethinkingInceptionArchitecture2016} &
   &
   &
   &
  0.701 &
  0.808 &
  0.402 &
  0.608 &
  0.791 &
  0.821 &
  0.353 &
  0.638 \\ \hline
DRF - GS &
  0.265 &
  0.439 &
  0.224 &
  0.558 &
  0.794 &
  0.469 &
  0.650 &
  0.539 &
  0.757 &
  0.353 &
  0.515 \\
DRF - ILP &
  0.193 &
  0.278 &
  0.306 &
  0.403 &
  0.672 &
  0.509 &
  0.647 &
  0.630 &
  0.734 &
  0.597 &
  0.727 \\
DRF - IJGP &
  0.312 &
  0.437 &
  {\ul 0.319} &
  0.561 &
  0.814 &
  0.439 &
  {\ul 0.701} &
  0.788 &
  0.853 &
  0.580 &
  0.737 \\ \hline
DDN - LR - Pipeline &
  0.345 &
  0.484 &
  0.290 &
  0.716 &
  0.826 &
  0.504 &
  0.672 &
  0.775 &
  0.855 &
  0.573 &
  0.702 \\
DDN - LR - Joint &
  {\ul 0.396} &
  0.548 &
  0.313 &
  {\ul 0.746} &
  0.839 &
  0.537 &
  0.686 &
  {\ul 0.843} &
  0.869 &
  {\ul 0.634} &
  {\ul 0.779} \\
DDN - MLP - Pipeline &
  0.375 &
  {\ul 0.549} &
  0.295 &
  0.729 &
  {\ul 0.860} &
  {\ul 0.579} &
  0.695 &
  0.812 &
  {\ul 0.882} &
  0.618 &
  0.727 \\
DDN - MLP - Joint &
  \textbf{0.407} &
  \textbf{0.554} &
  \textbf{0.341} &
  \textbf{0.780} &
  \textbf{0.875} &
  \textbf{0.596} &
  \textbf{0.704} &
  \textbf{0.881} &
  \textbf{0.897} &
  \textbf{0.697} &
  \textbf{0.792} \\ \hline
Relative Improvement (\%) &
  4.85 &
  3.66 &
  15.93 &
  11.20 &
  8.37 &
  48.43 &
  15.82 &
  11.34 &
  9.26 &
  97.66 &
  24.15 \\ \hline
\end{tabular}%
\end{sc}
}
\vspace{-4mm}
\end{table*}
\begin{table*}[t]
\centering
\caption{Comparison of our methods with the baseline for MLIC task. The best/second best values are bold/underlined. The last row shows the relative improvement made by the best-performing proposed method over the baseline.}
\resizebox{\textwidth}{!}{%
\begin{sc}
\begin{tabular}{c|cccc|cccc|cccc}
\hline
\multirow{2}{*}{Method} &
  \multicolumn{4}{c|}{MS-COCO} &
  \multicolumn{4}{c|}{NUS-WIDE} &
  \multicolumn{4}{c}{PASCAL-VOC} \\ \cline{2-13} 
 &
  mAP ↑ &
  LRAP ↑ &
  SA ↑ &
  JI ↑ &
  mAP ↑ &
  LRAP ↑ &
  SA ↑ &
  JI ↑ &
  mAP ↑ &
  LRAP ↑ &
  SA ↑ &
  JI ↑ \\ \hline
Q2L \cite{liuQuery2LabelSimpleTransformer2021} &
  \textbf{0.912} &
  \textbf{0.961} &
  0.507 &
  0.802 &
   &
   &
   &
   &
   &
   &
   &
   \\
MSRN \citep{quMultilayeredSemanticRepresentation2021} &
   &
   &
   &
   &
  \textbf{0.615} &
  {\ul 0.845} &
  0.314 &
  {\ul 0.638} &
  {\ul 0.960} &
  {\ul 0.976} &
  0.708 &
  0.853 \\ \hline
DRF - GS &
  0.751 &
  0.861 &
  0.347 &
  0.692 &
  0.401 &
  0.739 &
  0.282 &
  0.547 &
  0.767 &
  0.933 &
  0.727 &
  0.834 \\
DRF - ILP &
  0.735 &
  0.825 &
  0.545 &
  0.817 &
  0.252 &
  0.591 &
  0.322 &
  0.590 &
  0.809 &
  0.879 &
  0.761 &
  {\ul 0.876} \\
DRF - IJGP &
  0.741 &
  0.902 &
  0.546 &
  0.818 &
  0.410 &
  0.752 &
  {\ul 0.344} &
  0.628 &
  0.832 &
  0.941 &
  0.763 &
  0.869 \\ \hline
DDN - LR - Pipeline &
  0.830 &
  0.924 &
  0.496 &
  0.785 &
  0.432 &
  0.797 &
  0.306 &
  0.586 &
  0.884 &
  0.932 &
  0.684 &
  0.787 \\
DDN - LR - Joint &
  0.841 &
  0.928 &
  0.546 &
  0.816 &
  0.501 &
  0.821 &
  0.325 &
  0.623 &
  0.924 &
  0.962 &
  0.761 &
  0.869 \\
DDN - MLP - Pipeline &
  0.876 &
  0.945 &
  {\ul 0.556} &
  {\ul 0.821} &
  {\ul 0.561} &
  0.830 &
  0.332 &
  0.632 &
  0.927 &
  0.956 &
  {\ul 0.766} &
  {\ul 0.876} \\
DDN - MLP - Joint &
  {\ul 0.903} &
  {\ul 0.958} &
  \textbf{0.586} &
  \textbf{0.837} &
  \textbf{0.615} &
  \textbf{0.847} &
  \textbf{0.356} &
  \textbf{0.660} &
  \textbf{0.964} &
  \textbf{0.983} &
  \textbf{0.805} &
  \textbf{0.912} \\ \hline
Relative Improvement (\%) &
  -1.05 &
  -0.23 &
  15.44 &
  4.27 &
  0.00 &
  0.29 &
  13.42 &
  3.50 &
  0.37 &
  0.74 &
  13.70 &
  6.84 \\ \hline
\end{tabular}%
\end{sc}
}

\label{tab:eval-mlic}
\vskip -0.1in
\end{table*}

\icml{We compare the baseline neural networks  with three versions of DRFs and four versions of DDNs using the four metrics and six datasets given in Section \ref{sec:datasets}. The results are presented in tables \ref{tab:eval-mlac} and \ref{tab:eval-mlic}. We also show the improvements that our method makes over the baseline. Further evaluation on PASCAL-VOC and MS-COCO datasets can be found in the appendix, where we provide comparison between the proposed method and other state-of-the-art methods.}

\textbf{Comparison between Baseline neural network and DRFs}. \icml{We observe that IJGP and ILP outperform the baseline neural networks (which includes transformers for some datasets) in terms of the two non-probabilistic metrics JI and SA on five out of the six datasets. IJGP typically outperforms GS and ILP on JI. ILP outperforms the baseline on SA (notice that SA is 1 if there is an exact match between predicted and true labels and 0 otherwise) because it performs an accurate maximum-a-posteriori (MAP) inference (accurate MAP inference on an accurate model is likely to yield high SA). However, on metrics that require estimating the posterior probabilities, mAP and LRAP, the DRF schemes sometimes hurt the performance and at other times are only marginally better than the baseline methods. We observe that advanced inference schemes, particularly IJGP and ILP are superior on average to GS. Note that getting a higher SA is much harder in datasets having high label cardinalities. Specifically, SA does not distinguish between models that predict \textit{almost} correct labels and completely incorrect outputs.}

\textbf{Comparison between Baseline neural networks and DDNs}. We observe that the best performing DDN model, DDN-MLP with joint learning, outperforms the baseline neural networks on \icml{five out of the six datasets} on all metrics. Sometimes, the improvement is substantial (e.g., 9\% improvement in mAP on the wetlab dataset). \icml{The DDN-MLP model with joint learning improves considerably over the baseline method when performance is measured using SA and JI while keeping the precision comparative or higher.} Roughly speaking, the MLP versions are superior to the LR versions, and the joint models are superior to the ones trained using the pipeline method. On the non-probabilistic metrics (JI and SA), the pipeline models are often superior to the baseline neural network, while on the mAP metric, they may hurt the performance.

We observe that on the MLIC task, the DDN methods outperform Q2L and MSRN, even though both Q2L and MSRN model label correlations. This suggests that DDNs are either able to uncover additional relationships between labels during the learning phase or better reason about them during the inference (Gibbs sampling) phase or both. In particular, both Q2L and MSRN do not use Gibbs sampling to predict the labels, because they do not explicitly model the joint probability distribution over the labels.

\sva{Should we place the next line here or in the last paragraph of this section (summary)?}

\begin{figure*}[t!]
\begin{center}
\centerline{
  \includegraphics[width=\textwidth]{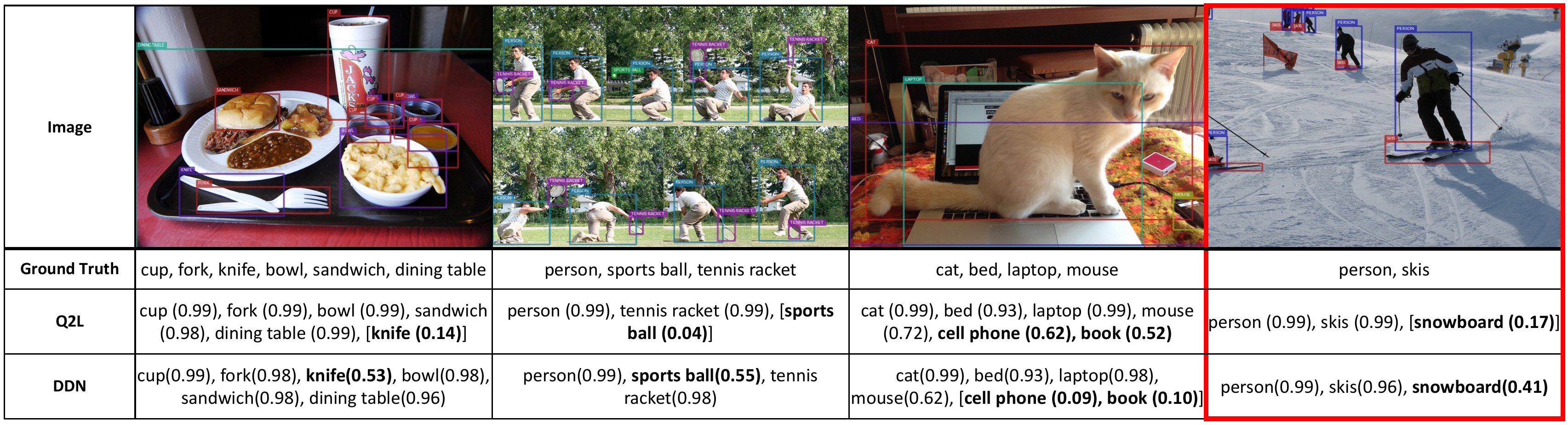}}
  \caption{Comparison of labels predicted by Q2L~\cite{liuQuery2LabelSimpleTransformer2021} and our DDN-MLP-Joint model on the MS-COCO dataset. Labels in bold represent the difference between the predictions of the two methods assuming that a threshold of 0.3 is used (i.e., every label whose probability $ > 0.3$ is considered as a predicted label). We also provide the probabilities in (). Labels enclosed in [] denote the labels that were not predicted by the corresponding method (added to compare the probabilities). The first three column shows examples where DDN improves over Q2L, while the last column (outlined in red) shows an example where DDN is worse than Q2L. }
  \label{tab:annotation_comparison}
  \end{center}
\vspace{-4mm}
\end{figure*}

\icml{In Figure \ref{tab:annotation_comparison}, we show a few images and their corresponding labels predicted using Q2L and DDN-MLP-Joint on the MS-COCO dataset. These results show that our method not only adds labels missed by Q2L but also removes several incorrect predictions. For example, in the first and second images, our method adds labels missed by Q2L and aligns the results perfectly with the ground truth. In the third image, our proposed method removes incorrect predictions. In the last image, we show an example where the DDN performs worse than Q2L and adds a label that is not in the ground truth. More examples are provided in the appendix.}

\textbf{Comparison between DRFs and DDNs}. We observe that the jointly trained DDNs are almost always superior to the best-performing DRFs on all datasets. Interestingly, on average, the pipeline DDN models outperform the DRF models when performance is measured using the mAP and LRAP metrics. However, when the SA and JI metrics are used, we observe that there is no significant difference in performance between pipeline DDNs and DRFs. \icml{Thus, DRFs can be especially beneficial if there are no GPU resource available for training and we want to optimize for JI or SA.}


In summary, jointly trained deep dependency networks are superior to the baseline neural networks as well as the models that combine Markov random fields and neural networks. The experimental results on MLC clearly demonstrate the practical usefulness of our proposed method.

\section{Related Work}
\label{section:related_work}
A large number of methods have been proposed that train PGMs and NNs jointly. For example,  \citep{zhengConditionalRandomFields2015} proposed to combine conditional random fields (CRFs) and recurrent neural networks (RNNs), \citep{schwingFullyConnectedDeep2015, larssonMaxMarginLearningDeep2017, larssonProjectedGradientDescent2018, arnabHigherOrderConditional2016} showed how to combine CNNs and CRFs,  \citep{chenLearningDeepStructured2015} proposed to use densely connected graphical models with CNNs, and \citep{johnsonComposingGraphicalModels2017} combined latent graphical models with neural networks. As far as we know, ours is the first work that shows how to jointly train a dependency network, neural network hybrid. Another virtue of DDNs is that they are easy to train and parallelizable, making them an attractive choice.

\eat{The combination of PGMs and NNs has been applied to improve performance on a wide variety of real-world tasks. Notable examples include Semantic Segmentation \citep{arnabConditionalRandomFields2018, guoSemanticImageSegmentation2021}, human pose estimation \citep{tompsonJointTrainingConvolutional2014, liangLimbBasedGraphicalModel2018, songThinSlicingNetworkDeep2017, yangEndtoEndLearningDeformable2016}, semantic labeling of body parts \citep{kirillovJointTrainingGeneric2016}, stereo estimation \cite{knobelreiterEndtoEndTrainingHybrid2017}, language understanding \citep{yaoRecurrentConditionalRandom2014}, joint intent detection and slot filling \citep{xuConvolutionalNeuralNetwork2013}, 2D Hand-pose Estimation \citep{kongAdaptiveGraphicalModel2019}, depth estimation from a single monocular image \citep{liuDeepConvolutionalNeural2014}, Polyphonic Piano Music Transcription \citep{sigtiaEndtoEndNeuralNetwork2016a}, face sketch synthesis \cite{zhuLearningDeepPatch2021}, Sea Ice Floe Segmentation \citep{nagiRUFEffectiveSea2021} and Crowdsourcing aggregation \citep{liCrowdsourcingAggregationDeep2021}). As far as we know, ours is the first work that uses jointly trained PGM+NN combinations to solve the multi-label action classification task in videos \icml{and multi-label classification task in images}.}

\icml{The combination of PGMs and NNs has been applied to improve performance on a wide variety of real-world tasks. Notable examples include human pose estimation \citep{tompsonJointTrainingConvolutional2014, liangLimbBasedGraphicalModel2018, songThinSlicingNetworkDeep2017, yangEndtoEndLearningDeformable2016}, semantic labeling of body parts \citep{kirillovJointTrainingGeneric2016}, stereo estimation \cite{knobelreiterEndtoEndTrainingHybrid2017}, language understanding \citep{yaoRecurrentConditionalRandom2014}, joint intent detection and slot filling \citep{xuConvolutionalNeuralNetwork2013}, polyphonic piano music transcription \citep{sigtiaEndtoEndNeuralNetwork2016a}, face sketch synthesis \cite{zhuLearningDeepPatch2021}, sea ice floe segmentation \citep{nagiRUFEffectiveSea2021} and crowd-sourcing aggregation \citep{liCrowdsourcingAggregationDeep2021}). \sva{Added this paragraph from introduction}
These hybrid models have also been used for solving a range of computer vision tasks such as semantic segmentation \citep{arnabConditionalRandomFields2018, guoSemanticImageSegmentation2021}, image crowd counting \citep{hanImageCrowdCounting2017}, \eat{Not a CV task structured output prediction \citep{pmlr-doNeuralConditionalRandom2010},} visual relationship detection \citep{yuProbabilisticGraphicalModel_2022_CVPR}, modeling for epileptic seizure detection in multichannel EEG \citep{craleyIntegratingConvolutionalNeural2019a}, \eat{Not a CV task phonetic recognition \citep{morrisConditionalRandomFields2008},} face sketch synthesis \citep{zhangNeuralProbabilisticGraphical2020}, semantic image segmentation \citep{chenDeepLabSemanticImage2018, linEfficientPiecewiseTraining2015}, 2D Hand-pose Estimation \citep{kongAdaptiveGraphicalModel2019}, depth estimation from a single monocular image \citep{liuDeepConvolutionalNeural2014}, animal pose tracking \citep{wuDeepGraphPoseNEURIPS2020} and pose estimation \citep{chenArticulatedPoseEstimation2014}. \eat{Not a CV task and adaptive team training \citep{rollArtificialIntelligenceEducation2021}}} As far as we know, ours is the first work that uses jointly trained PGM+NN combinations to solve multi-label action (in videos) and image classification tasks.

To date, dependency networks have been used to solve various tasks such as collective classification \citep{Neville2003CollectiveCW}, binary classification \citep{gamez2006dependency, gamez2008robust}, multi-label classification  \citep{guo2011multi}, part-of-speech tagging \citep{tarantola2002dependency}, relation prediction \citep{katzourisInductiveLogicProgramming2022b} and collaborative filtering \citep{heckerman2000dependency}. Ours is the first work that combines DNs with sophisticated feature representations and performs joint training over these representations.

\section{Conclusion and Future Work}
\label{sec:conclusion}


\icml{More and more state-of-the-art methods for challenging applications of computer vision tasks usually use deep neural networks. Deep neural networks are good at extracting features in vision tasks like image classification, video classification, object detection, image segmentation, and others. Nevertheless, for more complex tasks involving multi-label classification, these methods cannot model crucial information like inter-label dependencies. In this paper, we proposed a new modeling framework called deep dependency networks (DDNs) that combines a dependency network with a neural network and demonstrated via experiments, on three video and three image datasets, that it outperforms the baseline neural network, sometimes by a substantial margin. The key advantage of DDNs is that they explicitly model and reason about the relationship between the labels, and often improve model performance without considerable overhead. DDNs are also able to model additional relationships that are missed by other state-of-the-art methods that use transformers, attention module, and GAT.}
In particular, DDNs are simple to use, admit fast learning and inference, are easy to parallelize, and can leverage modern GPU architectures.

Avenues for future work include: applying the setup described in the paper to other multi-label classification tasks in computer vision, natural language understanding, and speech recognition; developing advanced inference schemes for dependency networks; converting DDNs to MRFs for better inference \citep{lowdClosedFormLearningMarkov2012a}; etc.


\eat{The framework provided here can be used as a feature processor for any multi-label classification task, providing potentially significant improvements over the baseline CNNs with small added overhead for training and inference. The method also does not depend on a particular feature extractor, allowing for the use of the current state-of-the-art methods with no modification. }


\subsection*{Acknowledgements}
This work was supported in part by the DARPA Perceptually enabled Task Guidance (PTG) Program under contract number HR00112220005 and by the National Science Foundation CAREER award IIS-1652835.


\bibliographystyle{unsrtnat}
\bibliography{main}

\newpage
\appendix
\onecolumn


\section{Details on Inference for DDNs}
\begin{algorithm}[h]
   \caption{Inference Algorithm for DDNs}
   \label{alg:gibbs}
\begin{algorithmic}[1]
    \INPUT{video segment/image $\rvv$, number of sample $N$, DDN $\langle \mathcal{N},\mathcal{D} \rangle$}
    \OUTPUT{An estimate of marginal probability distribution over each label $X_i$ of the DDN given $\rvv$}
    \STATE $\rve = \mathbb{N}(\rvv)$
   \STATE Randomly initialize $\rmX=\rvx^{(0)}$.
   \FOR{$j=1$ {\bfseries to} $N$}
       \STATE $\pi \gets$ Generate random permutation of $[1,n]$.
       \FOR{$i=1$ {\bfseries to} $n$}
           \STATE $x_{\pi(i)}^{(j)} \sim P_{\pi(i)}(x_{\pi(i)} | \rvx_{\pi(1):\pi(i-1)}^{(j)}, \rvx_{\pi(i+1):\pi(n)}^{(j-1)}, \rve)$    
        \ENDFOR
   \ENDFOR
   \FOR{$i=1$ {\bfseries to} $n$}
   \STATE $\hat{P}_i\left(x_{i}|\rvv\right)=\frac{1}{N} \sum_{j=1}^{N} P_i\left(x_{i} \mid \rvx_{-i}^{(j)},\rve\right)$
   \ENDFOR 
   \STATE {\bfseries return} $ \left\{\hat{P}_i\left(x_i|\rvv\right)|i \in \{1,\ldots,n\} \right \}$
\end{algorithmic}
\end{algorithm}

In this section, we describe our inference procedure for DDNs (see Algorithm \ref{alg:gibbs}). The inputs to the algorithm are (1) a video segment/image $\rvv$, (2) the number of samples $N$ and (3) trained DDN model $\langle \mathcal{N}, \mathcal{D}\rangle$. The algorithm begins (see step 1) by extracting features $\rve$ from the video segment/image $\rvv$ by sending the latter through the neural network $\mathcal{N}$ (which represents the function $\mathbb{N}$). Then in steps 2--8, it generates $N$ samples via Gibbs sampling. The Gibbs sampling procedure begins with a random assignment to all the labels (step 2). Then at each iteration (steps 3--8), it first generates a random permutation $\pi$ over the $n$ labels and samples the labels one by one along the order $\pi$ (steps 5--7). To sample a label indexed by $\pi(i)$ at iteration $j$, we  compute $P_{\pi(i)}(x_{\pi(i)} | \rvx_{\pi(1):\pi(i-1)}^{(j)}, \rvx_{\pi(i+1):\pi(n)}^{(j-1)}, \rve)$ from the DN $\mathcal{D}$ where $\rvx_{\pi(1):\pi(i-1)}^{(j)}$ and $\rvx_{\pi(i+1):\pi(n)}^{(j-1)}$ denote the assignments to all labels ordered before $x_{\pi(i)}$ at iteration $j$ and the assignments to all labels ordered after $x_{\pi(i)}$ at iteration $j-1$ respectively.

After $N$ samples are generated via Gibbs sampling, the algorithm uses them to estimate (see steps 9--11) the (posterior) marginal probability distribution at each label $X_i$ given $\rvv$ using the mixture estimator \cite{liubook}. The algorithm terminates (see step 12) by returning these posterior estimates.

\eat{the end of this procedure we have N samples and now we need to estimate the marginal probability
distribution of each label $X_i$ which we do by performing the mixture estimator given in \eqref{eq:Mixture_Estimator1}. Finally the marginal probability distribution is returned for all the labels.}

\section{Additional Evaluations for the MLIC task}
\icml{For the image classification task we report additional metrics other than the ones reported in section \ref{sec:experiments}. These metrics are usually used for the comparison of state of the art methods for the MLIC task and we report \emph{per-class average precision scores} for the PASCAL-VOC dataset and various \emph{top-one} and \emph{top-three scores} for the MS-COCO dataset.}
\subsection{PASCAL-VOC 2007}

\icml{We report the Average Precision scores for each class for the PASCAL-VOC dataset. The comparison is made between our best-performing method (DDN - MLP - Joint) and previous state-of-the-art methods including CNN-RNN \citep{wangCNNRNNUnifiedFramework2016}, VGG+SVM \citep{simonyanVeryDeepConvolutional2014}, Fev+Lv \citep{yangExploitBoundingBox2016}, HCP \citep{weiHCPFlexibleCNN2016}, RDAL \citep{wangMultilabelImageRecognition2017b}, RARL \citep{chenRecurrentAttentionalReinforcement2018}, SSGRL \citep{chenLearningSemanticSpecificGraph2019a}, MCAR \citep{gaoLearningDiscoverMultiClass2021}, ASL(TResNetL) \citep{ridnikAsymmetricLossMultiLabel2021}, ADD-GCN  \citep{yeAttentionDrivenDynamicGraph2020}, Q2L-TResL \citep{liuQuery2LabelSimpleTransformer2021}, ResNet-101 \citep{heDeepResidualLearning2016}, ML-GCN \citep{chenMultiLabelImageRecognition2019a}, and MSRN \citep{quMultilayeredSemanticRepresentation2021}. 

\begingroup

\setlength{\tabcolsep}{2.5pt} 
\renewcommand{\arraystretch}{1.1} 

\begin{table}[h]
\centering
\caption{Comparison of mAP and AP (in \%) of our method and state-of-the-art methods on Pascal VOC2007 dataset. Numbers in bold indicate the best performance}
\label{tab:voc_ap}
\resizebox{\textwidth}{!}{%
\begin{sc}
\begin{tabular}{c|cccccccccccccccccccc|c}
\hline
{\color[HTML]{212529} Methods} & {\color[HTML]{212529} aero} & {\color[HTML]{212529} bike} & {\color[HTML]{212529} bird} & {\color[HTML]{212529} \textbf{boat}} & {\color[HTML]{212529} \textbf{bottle}} & {\color[HTML]{212529} \textbf{bus}} & {\color[HTML]{212529} \textbf{car}} & {\color[HTML]{212529} \textbf{cat}} & {\color[HTML]{212529} \textbf{chair}} & {\color[HTML]{212529} \textbf{cow}} & {\color[HTML]{212529} \textbf{table}} & {\color[HTML]{212529} \textbf{dog}} & {\color[HTML]{212529} \textbf{horse}} & {\color[HTML]{212529} \textbf{mbike}} & {\color[HTML]{212529} \textbf{person}} & {\color[HTML]{212529} \textbf{plant}} & {\color[HTML]{212529} \textbf{sheep}} & {\color[HTML]{212529} \textbf{sofa}} & {\color[HTML]{212529} \textbf{train}} & {\color[HTML]{212529} \textbf{tv}} & {\color[HTML]{212529} \textbf{mAP}} \\ \hline
{\color[HTML]{212529} \textbf{CNN-RNN}} & {\color[HTML]{212529} 96.7} & {\color[HTML]{212529} 83.1} & {\color[HTML]{212529} 94.2} & {\color[HTML]{212529} 92.8} & {\color[HTML]{212529} 61.2} & {\color[HTML]{212529} 82.1} & {\color[HTML]{212529} 89.1} & {\color[HTML]{212529} 94.2} & {\color[HTML]{212529} 64.2} & {\color[HTML]{212529} 83.6} & {\color[HTML]{212529} 70} & {\color[HTML]{212529} 92.4} & {\color[HTML]{212529} 91.7} & {\color[HTML]{212529} 84.2} & {\color[HTML]{212529} 93.7} & {\color[HTML]{212529} 59.8} & {\color[HTML]{212529} 93.2} & {\color[HTML]{212529} 75.3} & {\color[HTML]{212529} 99.7} & {\color[HTML]{212529} 78.6} & {\color[HTML]{212529} 84} \\
{\color[HTML]{212529} \textbf{VGG+SVM}} & {\color[HTML]{212529} 98.9} & {\color[HTML]{212529} 95} & {\color[HTML]{212529} 96.8} & {\color[HTML]{212529} 95.4} & {\color[HTML]{212529} 69.7} & {\color[HTML]{212529} 90.4} & {\color[HTML]{212529} 93.5} & {\color[HTML]{212529} 96} & {\color[HTML]{212529} 74.2} & {\color[HTML]{212529} 86.6} & {\color[HTML]{212529} 87.8} & {\color[HTML]{212529} 96} & {\color[HTML]{212529} 96.3} & {\color[HTML]{212529} 93.1} & {\color[HTML]{212529} 97.2} & {\color[HTML]{212529} 70} & {\color[HTML]{212529} 92.1} & {\color[HTML]{212529} 80.3} & {\color[HTML]{212529} 98.1} & {\color[HTML]{212529} 87} & {\color[HTML]{212529} 89.7} \\
{\color[HTML]{212529} \textbf{Fev+Lv}} & {\color[HTML]{212529} 97.9} & {\color[HTML]{212529} 97} & {\color[HTML]{212529} 96.6} & {\color[HTML]{212529} 94.6} & {\color[HTML]{212529} 73.6} & {\color[HTML]{212529} 93.9} & {\color[HTML]{212529} 96.5} & {\color[HTML]{212529} 95.5} & {\color[HTML]{212529} 73.7} & {\color[HTML]{212529} 90.3} & {\color[HTML]{212529} 82.8} & {\color[HTML]{212529} 95.4} & {\color[HTML]{212529} 97.7} & {\color[HTML]{212529} 95.9} & {\color[HTML]{212529} 98.6} & {\color[HTML]{212529} 77.6} & {\color[HTML]{212529} 88.7} & {\color[HTML]{212529} 78} & {\color[HTML]{212529} 98.3} & {\color[HTML]{212529} 89} & {\color[HTML]{212529} 90.6} \\
{\color[HTML]{212529} \textbf{HCP}} & {\color[HTML]{212529} 98.6} & {\color[HTML]{212529} 97.1} & {\color[HTML]{212529} 98} & {\color[HTML]{212529} 95.6} & {\color[HTML]{212529} 75.3} & {\color[HTML]{212529} 94.7} & {\color[HTML]{212529} 95.8} & {\color[HTML]{212529} 97.3} & {\color[HTML]{212529} 73.1} & {\color[HTML]{212529} 90.2} & {\color[HTML]{212529} 80} & {\color[HTML]{212529} 97.3} & {\color[HTML]{212529} 96.1} & {\color[HTML]{212529} 94.9} & {\color[HTML]{212529} 96.3} & {\color[HTML]{212529} 78.3} & {\color[HTML]{212529} 94.7} & {\color[HTML]{212529} 76.2} & {\color[HTML]{212529} 97.9} & {\color[HTML]{212529} 91.5} & {\color[HTML]{212529} 90.9} \\
{\color[HTML]{212529} \textbf{RDAL}} & {\color[HTML]{212529} 98.6} & {\color[HTML]{212529} 97.4} & {\color[HTML]{212529} 96.3} & {\color[HTML]{212529} 96.2} & {\color[HTML]{212529} 75.2} & {\color[HTML]{212529} 92.4} & {\color[HTML]{212529} 96.5} & {\color[HTML]{212529} 97.1} & {\color[HTML]{212529} 76.5} & {\color[HTML]{212529} 92} & {\color[HTML]{212529} 87.7} & {\color[HTML]{212529} 96.8} & {\color[HTML]{212529} 97.5} & {\color[HTML]{212529} 93.8} & {\color[HTML]{212529} 98.5} & {\color[HTML]{212529} 81.6} & {\color[HTML]{212529} 93.7} & {\color[HTML]{212529} 82.8} & {\color[HTML]{212529} 98.6} & {\color[HTML]{212529} 89.3} & {\color[HTML]{212529} 91.9} \\
{\color[HTML]{212529} \textbf{RARL}} & {\color[HTML]{212529} 98.6} & {\color[HTML]{212529} 97.1} & {\color[HTML]{212529} 97.1} & {\color[HTML]{212529} 95.5} & {\color[HTML]{212529} 75.6} & {\color[HTML]{212529} 92.8} & {\color[HTML]{212529} 96.8} & {\color[HTML]{212529} 97.3} & {\color[HTML]{212529} 78.3} & {\color[HTML]{212529} 92.2} & {\color[HTML]{212529} 87.6} & {\color[HTML]{212529} 96.9} & {\color[HTML]{212529} 96.5} & {\color[HTML]{212529} 93.6} & {\color[HTML]{212529} 98.5} & {\color[HTML]{212529} 81.6} & {\color[HTML]{212529} 93.1} & {\color[HTML]{212529} 83.2} & {\color[HTML]{212529} 98.5} & {\color[HTML]{212529} 89.3} & {\color[HTML]{212529} 92} \\
{\color[HTML]{212529} \textbf{SSGRL}} & {\color[HTML]{212529} 99.7} & {\color[HTML]{212529} 98.4} & {\color[HTML]{212529} 98} & {\color[HTML]{212529} 97.6} & {\color[HTML]{212529} 85.7} & {\color[HTML]{212529} 96.2} & {\color[HTML]{212529} 98.2} & {\color[HTML]{212529} 98.8} & {\color[HTML]{212529} 82} & {\color[HTML]{212529} 98.1} & {\color[HTML]{212529} \textbf{89.7}} & {\color[HTML]{212529} 98.8} & {\color[HTML]{212529} 98.7} & {\color[HTML]{212529} 97} & {\color[HTML]{212529} 99} & {\color[HTML]{212529} 86.9} & {\color[HTML]{212529} 98.1} & {\color[HTML]{212529} 85.8} & {\color[HTML]{212529} 99} & {\color[HTML]{212529} 93.7} & {\color[HTML]{212529} 95} \\
{\color[HTML]{212529} \textbf{MCAR}} & {\color[HTML]{212529} 99.7} & {\color[HTML]{212529} \textbf{99}} & {\color[HTML]{212529} 98.5} & {\color[HTML]{212529} 98.2} & {\color[HTML]{212529} 85.4} & {\color[HTML]{212529} 96.9} & {\color[HTML]{212529} 97.4} & {\color[HTML]{212529} 98.9} & {\color[HTML]{212529} 83.7} & {\color[HTML]{212529} 95.5} & {\color[HTML]{212529} 88.8} & {\color[HTML]{212529} 99.1} & {\color[HTML]{212529} 98.2} & {\color[HTML]{212529} 95.1} & {\color[HTML]{212529} 99.1} & {\color[HTML]{212529} 84.8} & {\color[HTML]{212529} 97.1} & {\color[HTML]{212529} 87.8} & {\color[HTML]{212529} 98.3} & {\color[HTML]{212529} 94.8} & {\color[HTML]{212529} 94.8} \\
{\color[HTML]{212529} \textbf{ASL(TResNetL)}} & {\color[HTML]{212529} 99.9} & {\color[HTML]{212529} 98.4} & {\color[HTML]{212529} 98.9} & {\color[HTML]{212529} 98.7} & {\color[HTML]{212529} 86.8} & {\color[HTML]{212529} 98.2} & {\color[HTML]{212529} 98.7} & {\color[HTML]{212529} 98.5} & {\color[HTML]{212529} 83.1} & {\color[HTML]{212529} 98.3} & {\color[HTML]{212529} 89.5} & {\color[HTML]{212529} 98.8} & {\color[HTML]{212529} 99.2} & {\color[HTML]{212529} 98.6} & {\color[HTML]{212529} 99.3} & {\color[HTML]{212529} 89.5} & {\color[HTML]{212529} 99.4} & {\color[HTML]{212529} 86.8} & {\color[HTML]{212529} 99.6} & {\color[HTML]{212529} 95.2} & {\color[HTML]{212529} 95.8} \\
{\color[HTML]{212529} \textbf{ADD-GCN}} & {\color[HTML]{212529} 99.8} & {\color[HTML]{212529} \textbf{99}} & {\color[HTML]{212529} 98.4} & {\color[HTML]{212529} 99} & {\color[HTML]{212529} 86.7} & {\color[HTML]{212529} 98.1} & {\color[HTML]{212529} 98.5} & {\color[HTML]{212529} 98.3} & {\color[HTML]{212529} 85.8} & {\color[HTML]{212529} 98.3} & {\color[HTML]{212529} 88.9} & {\color[HTML]{212529} 98.8} & {\color[HTML]{212529} 99} & {\color[HTML]{212529} 97.4} & {\color[HTML]{212529} 99.2} & {\color[HTML]{212529} 88.3} & {\color[HTML]{212529} 98.7} & {\color[HTML]{212529} \textbf{90.7}} & {\color[HTML]{212529} 99.5} & {\color[HTML]{212529} \textbf{97}} & {\color[HTML]{212529} 96} \\
{\color[HTML]{212529} \textbf{Q2L-TResL}} & {\color[HTML]{212529} 99.9} & {\color[HTML]{212529} 98.9} & {\color[HTML]{212529} 99} & {\color[HTML]{212529} 98.4} & {\color[HTML]{212529} 87.7} & {\color[HTML]{212529} \textbf{98.6}} & {\color[HTML]{212529} 98.8} & {\color[HTML]{212529} 99.1} & {\color[HTML]{212529} 84.5} & {\color[HTML]{212529} 98.3} & {\color[HTML]{212529} 89.2} & {\color[HTML]{212529} 99.2} & {\color[HTML]{212529} 99.2} & {\color[HTML]{212529} 99.2} & {\color[HTML]{212529} 99.3} & {\color[HTML]{212529} 90.2} & {\color[HTML]{212529} 98.8} & {\color[HTML]{212529} 88.3} & {\color[HTML]{212529} 99.5} & {\color[HTML]{212529} 95.5} & {\color[HTML]{212529} 96.1} \\
{\color[HTML]{212529} \textbf{ResNet-101}} & {\color[HTML]{212529} 99.5} & {\color[HTML]{212529} 97.7} & {\color[HTML]{212529} 97.8} & {\color[HTML]{212529} 96.4} & {\color[HTML]{212529} 75.7} & {\color[HTML]{212529} 91.8} & {\color[HTML]{212529} 96.1} & {\color[HTML]{212529} 97.6} & {\color[HTML]{212529} 74.2} & {\color[HTML]{212529} 80.9} & {\color[HTML]{212529} 85} & {\color[HTML]{212529} 98.4} & {\color[HTML]{212529} 96.5} & {\color[HTML]{212529} 95.9} & {\color[HTML]{212529} 98.4} & {\color[HTML]{212529} 70.1} & {\color[HTML]{212529} 88.3} & {\color[HTML]{212529} 80.2} & {\color[HTML]{212529} 98.9} & {\color[HTML]{212529} 89.2} & {\color[HTML]{212529} 89.9} \\
{\color[HTML]{212529} \textbf{ML-GCN}} & {\color[HTML]{212529} 99.5} & {\color[HTML]{212529} 98.5} & {\color[HTML]{212529} 98.6} & {\color[HTML]{212529} 98.1} & {\color[HTML]{212529} 80.8} & {\color[HTML]{212529} 94.6} & {\color[HTML]{212529} 97.2} & {\color[HTML]{212529} 98.2} & {\color[HTML]{212529} 82.3} & {\color[HTML]{212529} 95.7} & {\color[HTML]{212529} 86.4} & {\color[HTML]{212529} 98.2} & {\color[HTML]{212529} 98.4} & {\color[HTML]{212529} 96.7} & {\color[HTML]{212529} 99} & {\color[HTML]{212529} 84.7} & {\color[HTML]{212529} 96.7} & {\color[HTML]{212529} 84.3} & {\color[HTML]{212529} 98.9} & {\color[HTML]{212529} 93.7} & {\color[HTML]{212529} 94} \\
{\color[HTML]{212529} \textbf{MSRN}} & {\color[HTML]{212529} \textbf{100}} & {\color[HTML]{212529} 98.8} & {\color[HTML]{212529} 98.9} & {\color[HTML]{212529} 99.1} & {\color[HTML]{212529} 81.6} & {\color[HTML]{212529} 95.5} & {\color[HTML]{212529} 98} & {\color[HTML]{212529} 98.2} & {\color[HTML]{212529} 84.4} & {\color[HTML]{212529} 96.6} & {\color[HTML]{212529} 87.5} & {\color[HTML]{212529} 98.6} & {\color[HTML]{212529} 98.6} & {\color[HTML]{212529} 97.2} & {\color[HTML]{212529} 99.1} & {\color[HTML]{212529} 87} & {\color[HTML]{212529} 97.6} & {\color[HTML]{212529} 86.5} & {\color[HTML]{212529} 99.4} & {\color[HTML]{212529} 94.4} & {\color[HTML]{212529} 94.9} \\
{\color[HTML]{212529} \textbf{MRSN(pre)}} & {\color[HTML]{212529} 99.7} & {\color[HTML]{212529} 98.9} & {\color[HTML]{212529} 98.7} & {\color[HTML]{212529} 99.1} & {\color[HTML]{212529} 86.6} & {\color[HTML]{212529} 97.9} & {\color[HTML]{212529} 98.5} & {\color[HTML]{212529} 98.9} & {\color[HTML]{212529} 86} & {\color[HTML]{212529} 98.7} & {\color[HTML]{212529} 89.1} & {\color[HTML]{212529} 99} & {\color[HTML]{212529} 99.1} & {\color[HTML]{212529} 97.3} & {\color[HTML]{212529} 99.2} & {\color[HTML]{212529} 90.2} & {\color[HTML]{212529} 99.2} & {\color[HTML]{212529} 89.7} & {\color[HTML]{212529} 99.8} & {\color[HTML]{212529} 95.3} & {\color[HTML]{212529} 96} \\ \hline
{\color[HTML]{212529} \textbf{DDN (ours)}} & 99.9 & 96.5 & \textbf{99.9} & \textbf{99.9} & \textbf{97.3} & 98.2 & \textbf{99.2} & \textbf{99.9} & \textbf{87.1} & \textbf{99.5} & 87.8 & \textbf{99.7} & \textbf{99.3} & \textbf{99.4} & \textbf{99.4} & \textbf{92.8} & \textbf{99.9} & 75.6 & \textbf{99.9} & 96.7 & \textbf{96.4} \\ \hline
\end{tabular}%
\end{sc}
}
\end{table}
\endgroup

The results are presented in table \ref{tab:voc_ap}. The proposed method is better than the other methods on fourteen out of the twenty labels, suggesting that DDNs are able to better model and reason about the inter-label dependencies than other competing methods, some of which also try to model these relationships. We also get the \emph{highest mAP scores} among all the methods. Also, note that the metrics for the backbone (and baseline for the PASCAL-VOC dataset) are in the last but two rows. DDN-MLP-Joint outperforms the backbone on seventeen labels. This shows that using a DDN as a multi-label classification head on top of state-of-the-art methods can help improve results by a high margin without being computationally expensive.}

\subsection{MS-COCO}
Table \ref{tab:appendix_coco} presents additional evaluation metrics for the \emph{MS-COCO dataset}. Specifically, we report overall precision (OP), recall (OR), F1-measure (OF1) and, per-category precision (CP), recall (CR), and F1-measure (CF1) for all and top-3 predicted labels. Note that in literature, OF1 and CF1 are more commonly used as compared to the other metrics to evaluate models for MLIC. We compare our method with SRN \citep{zhuLearningSpatialRegularization2017}, CADM \citep{chenMultiLabelImageRecognition2019c} , ML-GCN \citep{chenMultiLabelImageRecognition2019a}, KSSNet \citep{MultiLabelImageClassification}, MS-CMA \citep{youCrossModalityAttentionSemantic2020}, MCAR \citep{gaoLearningDiscoverMultiClass2021}, SSGRL \cite{chenLearningSemanticSpecificGraph2019a}, C-Trans \citep{lanchantinGeneralMultilabelImage2021}, ADD-GCN \citep{yeAttentionDrivenDynamicGraph2020} , ASL \citep{ridnikAsymmetricLossMultiLabel2021}, MlTr-l \citep{chengMLTRMultiLabelClassification2022}, Swin-L \citep{liuSwinTransformerHierarchical2021}, CvT-w24 \citep{wuCvTIntroducingConvolutions2021} and Q2L-CvT \citep{liuQuery2LabelSimpleTransformer2021}. 

The main advantage our proposed method has over the methods mentioned above is that it can be utilized for any MLC task, while most of the methods mentioned above can only be used for MLIC task. As we show in the section \ref{sec:experiments}, DDNs can be applied to the task of MLAC as well. As long as a feature extractor can extract features from the data (of any modality: videos, natural language, speech, etc.), DDNs can be used to model and reason about the relationships between the labels.

\begin{table}[h]
\centering
\caption{Comparison of our method with state-of-the-art models on the MS-COCO dataset. Note that the OF1 and CF1 are the metrics used most commonly in the literature as they do not depend on any hyper-parameters.}
\label{tab:appendix_coco}
\resizebox{0.9\textwidth}{!}{%
\begin{sc}
\begin{tabular}{l|llllll|llllll}
\hline
\multicolumn{1}{c|}{\multirow{2}{*}{Method}} & \multicolumn{6}{c|}{All} & \multicolumn{6}{c}{Top 3} \\ \cline{2-13} 
\multicolumn{1}{c|}{} & CP & CR & CF1 & OP & OR & OF1 & CP & CR & CF1 & OP & OR & OF1 \\ \hline
SRN & 81.6 & 65.4 & 71.2 & 82.7 & 69.9 & 75.8 & 85.2 & 58.8 & 67.4 & 87.4 & 62.5 & 72.9 \\
CADM & 82.5 & 72.2 & 77 & 84 & 75.6 & 79.6 & 87.1 & 63.6 & 73.5 & 89.4 & 66 & 76 \\
ML-GCN & 85.1 & 72 & 78 & 85.8 & 75.4 & 80.3 & 87.2 & 64.6 & 74.2 & 89.1 & 66.7 & 76.3 \\
KSSNet & 84.6 & 73.2 & 77.2 & 87.8 & 76.2 & 81.5 & - & - & - & - & - & - \\
MS-CMA & 82.9 & 74.4 & 78.4 & 84.4 & 77.9 & 81 & 86.7 & 64.9 & 74.3 & 90.9 & 67.2 & 77.2 \\
MCAR & 85 & 72.1 & 78 & 88 & 73.9 & 80.3 & 88.1 & 65.5 & 75.1 & 91 & 66.3 & 76.7 \\
SSGRL & \textbf{89.9} & 68.5 & 76.8 & \textbf{91.3} & 70.8 & 79.7 & 91.9 & 62.5 & 72.7 & 93.8 & 64.1 & 76.2 \\
C-Trans & 86.3 & 74.3 & 79.9 & 87.7 & 76.5 & 81.7 & 90.1 & 65.7 & 76 & 92.1 & 71.4 & 77.6 \\
ADD-GCN & 84.7 & 75.9 & 80.1 & 84.9 & 79.4 & 82 & 88.8 & 66.2 & 75.8 & 90.3 & 68.5 & 77.9 \\
ASL & 87.2 & 76.4 & 81.4 & 88.2 & 79.2 & 81.8 & 91.8 & 63.4 & 75.1 & 92.9 & 66.4 & 77.4 \\
MlTr-l & 86 & 81.4 & 83.3 & 86.5 & 83.4 & 84.9 & - & - & - & - & - & - \\
Swin-L & \textbf{89.9} & 80.2 & 84.8 & 90.4 & 82.1 & 86.1 & \textbf{93.6} & 69.9 & 80 & 94.3 & 71.1 & 81.1 \\
CvT-w24 & 89.4 & 81.7 & 85.4 & 89.6 & 83.8 & 86.6 & 93.3 & 70.5 & 80.3 & 94.1 & 71.5 & 81.3 \\
Q2L-CvT & 88.8 & \textbf{83.2} & 85.9 & 89.2 & 84.6 & 86.8 & 92.8 & \textbf{71.6} & \textbf{80.8} & 93.9 & 72.1 & 81.6 \\ \hline
DDN (Ours) & 88.4 & 82.7 & \textbf{86.0} & 90.6 & \textbf{85.3} & \textbf{87.9} & 91.8 & 69.9 & 79.4 & \textbf{94.8} & \textbf{72.2} & \textbf{82.0} \\ \hline
\end{tabular}%
\end{sc}
}
\end{table}

\section{Comparing Training and Inference Time for MLAC task}
In this section we will look at the computational time requirements of the methods mentioned in this paper. We compare both, the time it requires to train a model given training data and also the time it requires to perform inference for a given example. 
\begin{table}[h]
\centering
\caption{Time Comparisons for the Proposed methods. Training Time is in Hours and Inference times is in seconds. The inference was performed on a CPU, while training was performed on a GPU. For each dataset, we first show the time it takes to train the model, and in the second row, we show the time it takes to perform inference for a single example.}
\label{tab:time}
\begin{tabular}{|c|cc|cc|cc|}
\hline
                    & \multicolumn{2}{c|}{\textbf{Tacos}}                       & \multicolumn{2}{c|}{\textbf{Wetlab}}                      & \multicolumn{2}{c|}{\textbf{Charades}}                    \\ \cline{2-7} 
\textbf{}           & \multicolumn{1}{c|}{\textbf{Train}}  & \textbf{Inference} & \multicolumn{1}{c|}{\textbf{Train}}  & \textbf{Inference} & \multicolumn{1}{c|}{\textbf{Train}}  & \textbf{Inference} \\ \hline
\textbf{\mrfgs}     & \multicolumn{1}{c|}{$\sim$ 5 hrs}    & $\sim$ 0.58 sec    & \multicolumn{1}{c|}{$\sim$ 6.5 hrs}  & $\sim$ 0.61 sec    & \multicolumn{1}{c|}{$\sim$ 8 hrs}    & $\sim$ 1.93 sec    \\ \hline
\textbf{\ilp}       & \multicolumn{1}{c|}{$\sim$ 5 hrs}    & $\sim$ 1.54 sec    & \multicolumn{1}{c|}{$\sim$ 6.5 hrs}  & $\sim$ 1.46 sec    & \multicolumn{1}{c|}{$\sim$ 9 hrs}    & $\sim$ 2.42 sec    \\ \hline
\textbf{\ijgp}      & \multicolumn{1}{c|}{$\sim$ 5 hrs}    & $\sim$ 2.31 sec    & \multicolumn{1}{c|}{$\sim$ 6.5 hrs}  & $\sim$ 2.15 sec    & \multicolumn{1}{c|}{$\sim$ 9 hrs}    & $\sim$ 5.8 sec     \\ \hline
\textbf{\dnlr}      & \multicolumn{1}{c|}{$\sim$ 1.5 hrs}  & $\sim$ 0.1 sec     & \multicolumn{1}{c|}{$\sim$ 2 hrs}    & $\sim$ 0.15 sec    & \multicolumn{1}{c|}{$\sim$ 3 hrs}    & $\sim$ 0.39 sec    \\ \hline
\textbf{\dnlrjoint} & \multicolumn{1}{c|}{$\sim$ 6 hrs}    & $\sim$ 0.1 sec     & \multicolumn{1}{c|}{$\sim$ 7.25 hrs} & $\sim$ 0.15 sec    & \multicolumn{1}{c|}{$\sim$ 12 hrs}   & $\sim$ 0.39 sec    \\ \hline
\textbf{\dnnn}      & \multicolumn{1}{c|}{$\sim$ 2.25 hrs} & $\sim$ 0.19 sec    & \multicolumn{1}{c|}{$\sim$ 3 hrs}    & $\sim$ 0.31 sec    & \multicolumn{1}{c|}{$\sim$ 4.25 hrs} & $\sim$ 0.58 sec    \\ \hline
\textbf{\dnnnjoint} & \multicolumn{1}{c|}{$\sim$ 7 hrs}    & $\sim$ 0.19 sec    & \multicolumn{1}{c|}{$\sim$ 8 hrs}    & $\sim$ 0.31 sec    & \multicolumn{1}{c|}{$\sim$ 14.5 hrs} & $\sim$ 0.58 sec    \\ \hline
\end{tabular}
\end{table}
Let us compare different methods based on the information given in table \ref{tab:time}.

\textbf{Comparison among DRFs} \\
The learning time for the DRFs remains the same across the different methods, because we use the same models and apply different inference techniques on them. But for inference time, we can see that as we use more sophisticated methods, the inference time goes up. 

\textbf{Comparison among DDNs} \\
For DDNs the inference times remain the same for both, pipeline and joint models.  But for learning we see that joint model takes more time than the pipeline model. This is due to the fact that we are jointly learning the NN and the DN, and the inclusion of NNs for learning drives the time up.  

\textbf{Comparison between DRFs and DDNs} \\
The learning time for pipeline DDN models are significantly less than that of the DRF model. Both the pipeline model and the joint model are exceptionally faster than the DRF methods. These two observations  confirm our comments that DDNs are very fast and thus can be used in real time.

\section{Annotations comparison between Q2L and DDN-MLP-Joint on the MS-COCO dataset}
\icml{
In table \ref{tab:annotation_comparison_2}, we show more qualitative results which show the labels predicted by the best-performing proposed method (DDN-MLP-Joint) and the baseline method (Q2L) on various images from the MS-COCO dataset. This helps us to understand how and why the proposed method is able to achieve better evaluation scores, especially for subset accuracy, than the baseline.

In the first six rows, DDN adds labels to the baseline and gives correct results, while in the next ten examples, DDN removes the labels that Q2L predicted and predicts the exact same labels as ground truth. In the remaining examples, we look at cases where DDN adds/removes labels and yields incorrect predictions while the predictions from Q2L are correct. Note that the DDN is better than Q2L if we consider subset accuracy and its predictions are usually the same as the ground truth.}

\begin{longtable}{|p{0.27\textwidth}|p{0.1\textwidth}|p{0.28\textwidth}|p{0.28\textwidth}|}
\caption{Comparison of annotations produced by Q2L and DDN-MLP on the MS-COCO dataset. Labels in bold represent the difference between the two methods assuming that the threshold of 0.3 is used (i.e., every label whose probability is greater than 0.3 is considered a predicted label for that image). Values inside ( ) represent the probabilities of the corresponding predictions. Labels in [ ] represent labels that were not predicted by the corresponding method (Added to compare the probabilities).}
\label{tab:annotation_comparison_2}\\
\hline
\textbf{Image} & \textbf{Ground Truth} & \textbf{Q2L} & \textbf{DDN} \\ \hline
\endfirsthead
\endhead
\parbox[c]{1em}{\includegraphics[width=0.25\textwidth]{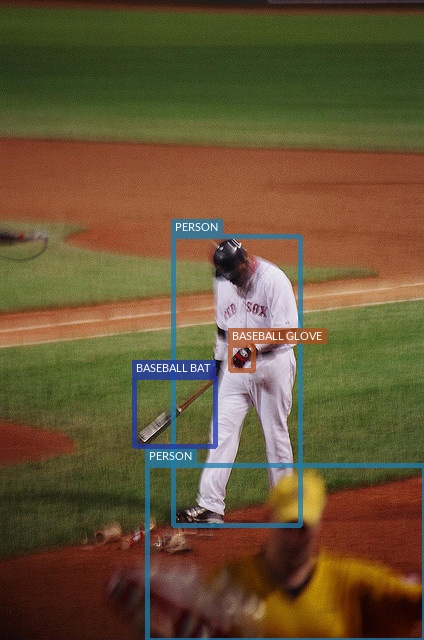}} & person, baseball bat, baseball   glove & person (0.99), baseball bat   (0.99), [\textbf{baseball glove (0.18)}] & person(0.99), baseball   bat(0.96), \textbf{baseball glove(0.44)} \\ \hline
\parbox[c]{1em}{\includegraphics[width=0.25\textwidth]{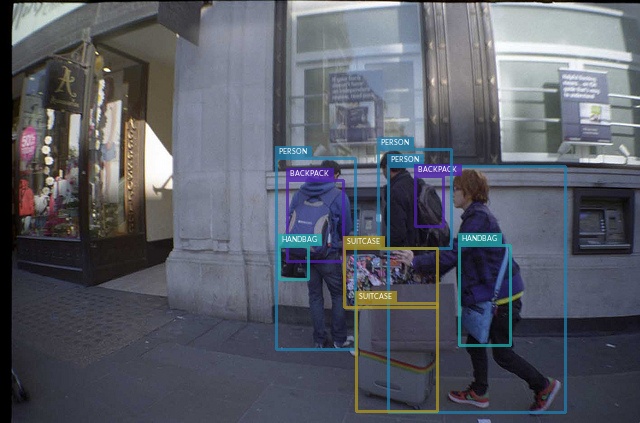}} & person, backpack, handbag,   suitcase & person (0.99), backpack (0.97),   handbag (0.99), [\textbf{suitcase (0.14)}] & person(0.99), backpack(0.96),   handbag(0.97), \textbf{suitcase(0.45)} \\ \hline
\parbox[c]{1em}{\includegraphics[width=0.25\textwidth]{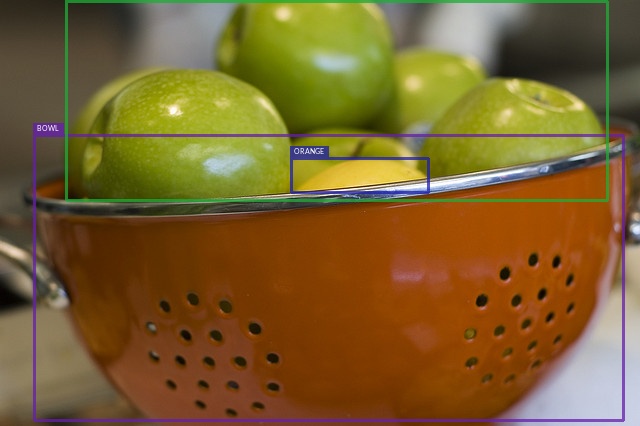}} & bowl, apple, orange & bowl (0.99), apple (0.99),   [\textbf{orange (0.03)}] & bowl(0.97), apple(0.96),  \textbf{orange(0.63)} \\ \hline
\parbox[c]{1em}{\includegraphics[width=0.25\textwidth]{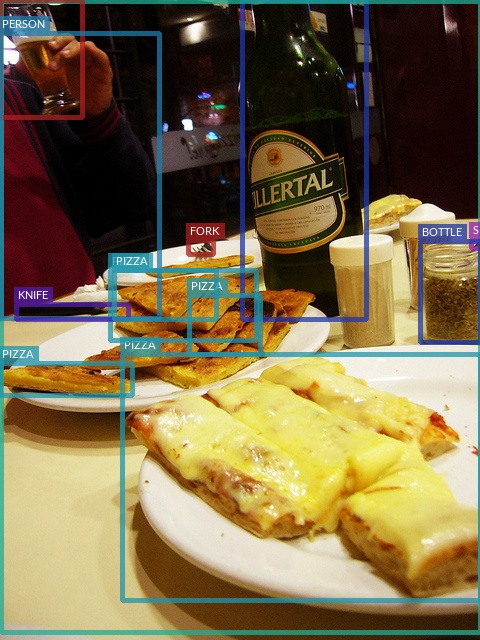}} & person, bottle, cup, fork,   knife, spoon, pizza, dining table & person (0.99), bottle (0.99),   cup (0.99), fork (0.99), knife (0.99), pizza (0.99), dining table (0.98),   [\textbf{spoon (0.20)}] & person(0.99), bottle(0.99),   cup(0.99), fork(0.98), knife(0.97), \textbf{spoon(0.72)}, pizza(0.99), dining   table(0.96) \\ \hline
\parbox[c]{1em}{\includegraphics[width=0.25\textwidth]{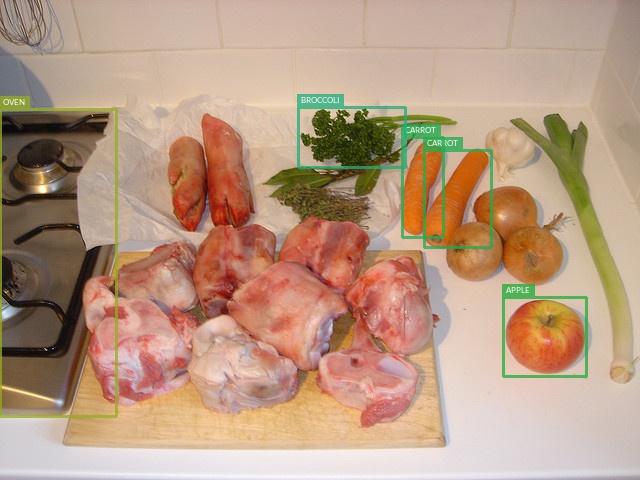}} & apple, broccoli, carrot, oven & apple (0.99), carrot (0.99),   oven (0.98), [\textbf{broccoli (0.20)}] & apple(0.94), \textbf{broccoli(0.78)},   carrot(0.95), oven(0.89) \\ \hline
\parbox[c]{1em}{\includegraphics[width=0.25\textwidth]{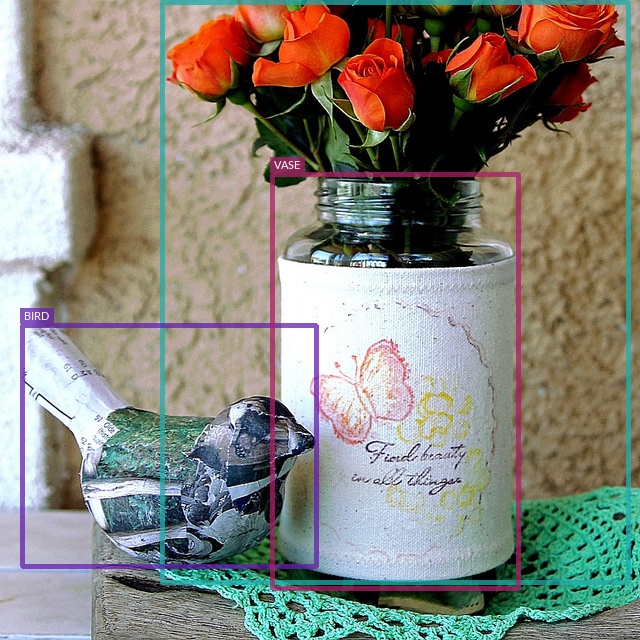}} & bird, potted plant, vase & vase (0.99), [\textbf{bird (0.08),   potted plant (0.28)}] & \textbf{bird(0.35), potted plant(0.74)},   vase(0.96) \\ \hline
\parbox[c]{1em}{\includegraphics[width=0.25\textwidth]{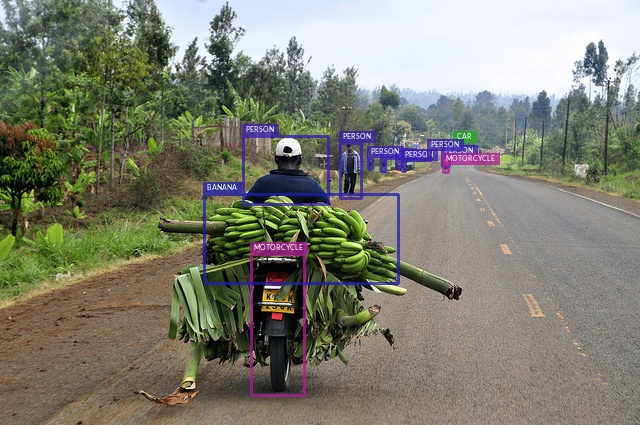}} & person, car, motorcycle,   banana & person (0.99), car (0.73),   motorcycle (0.99), \textbf{truck (0.48), cow (0.57)}, banana (0.99) & person(0.99), car(0.64),   motorcycle(0.99), banana(0.96), [\textbf{truck (0.16), cow (0.01)}] \\ \hline
\parbox[c]{1em}{\includegraphics[width=0.25\textwidth]{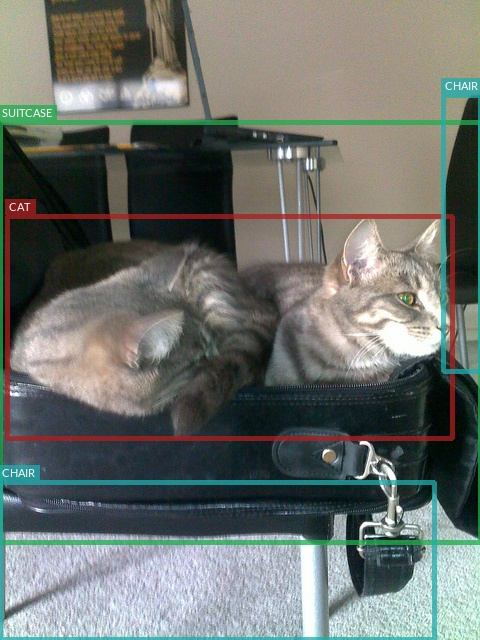}} & cat, suitcase, chair & cat (0.99), \textbf{handbag (0.45)},   suitcase (0.77), chair (0.99), \textbf{dining table (0.52)} & cat(0.98), suitcase(0.90),   chair(0.98), [\textbf{handbag (0.23), dining table (0.15)}] \\ \hline
\parbox[c]{1em}{\includegraphics[width=0.25\textwidth]{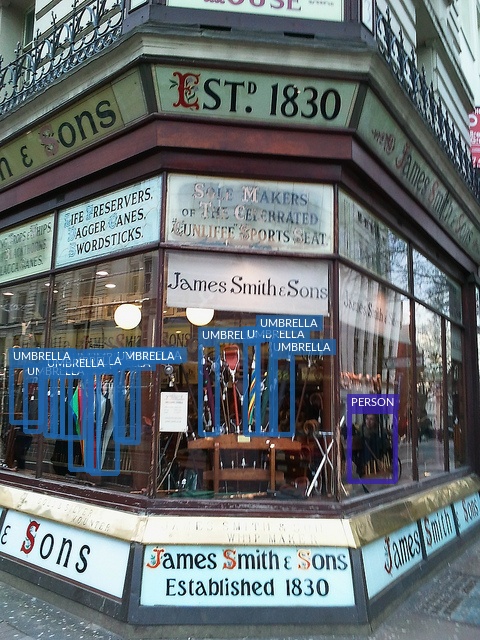}} & person, umbrella & person (0.97), \textbf{bicycle (0.44)},   umbrella (0.62), \textbf{tie (0.52)} & person(0.98), umbrella(0.61),   [\textbf{bicycle (0.09), tie (0.06)}] \\ \hline
\parbox[c]{1em}{\includegraphics[width=0.25\textwidth]{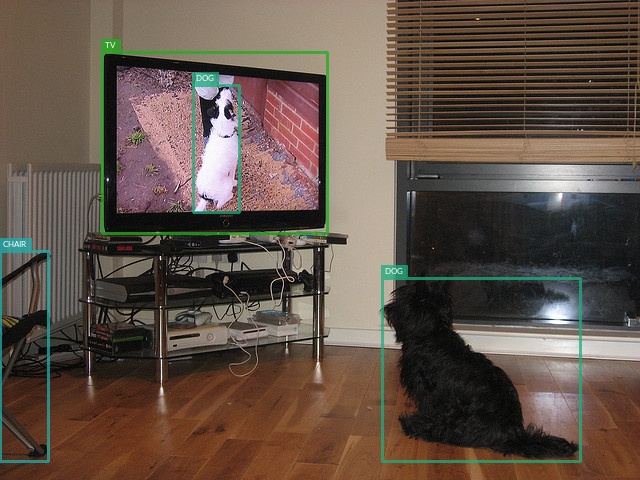}} & dog, chair, tv & \textbf{cat (0.69)}, dog (0.99), chair (0.98), tv (0.99), \textbf{book (0.42)} & dog(0.98), chair(0.97),   tv(0.98), [\textbf{cat (0.15), book (0.05)}] \\ \hline
\parbox[c]{1em}{\includegraphics[width=0.25\textwidth]{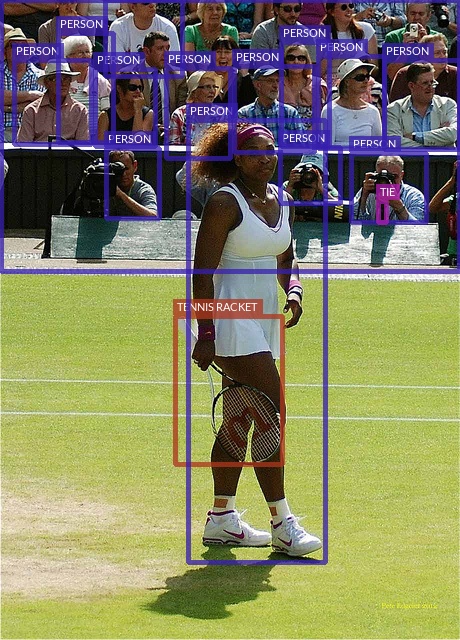}} & person, tie, tennis racket & person (0.99), \textbf{backpack (0.53),   handbag (0.64)}, tie (0.53), tennis racket (0.99) & person(0.99), tie(0.58), tennis   racket(0.96), [\textbf{backpack (0.07), handbag (0.24)}] \\ \hline
\parbox[c]{1em}{\includegraphics[width=0.25\textwidth]{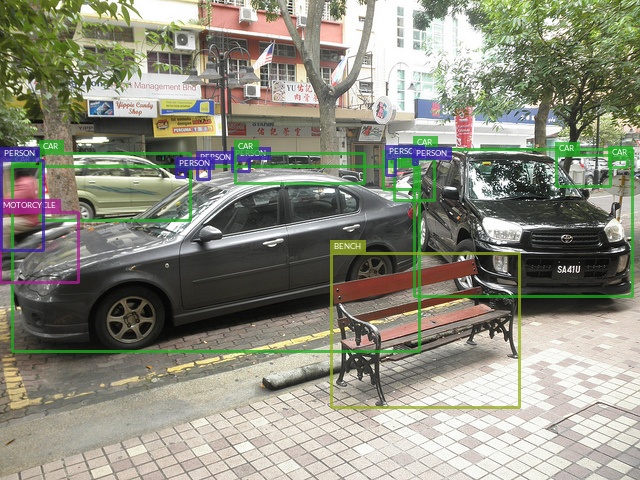}} & person, car, motorcycle, bench & person (0.99), car (0.99),   motorcycle (0.55), \textbf{parking meter (0.40)}, bench (0.99), \textbf{handbag (0.69)} & person(0.99), car(0.99),   motorcycle(0.86), bench(0.98), [\textbf{parking meter (0.09), handbag (0.16)}] \\ \hline
\parbox[c]{1em}{\includegraphics[width=0.25\textwidth]{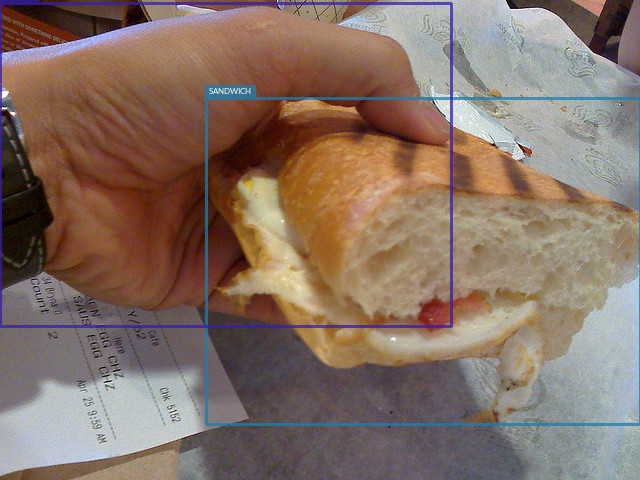}} & person, sandwich & person (0.99), sandwich (0.99), \textbf{cell phone (0.64), clock (0.43)} & person(0.99), sandwich(0.96),   [\textbf{cell phone (0.15), clock (0.07)}] \\ \hline
\parbox[c]{1em}{\includegraphics[width=0.25\textwidth]{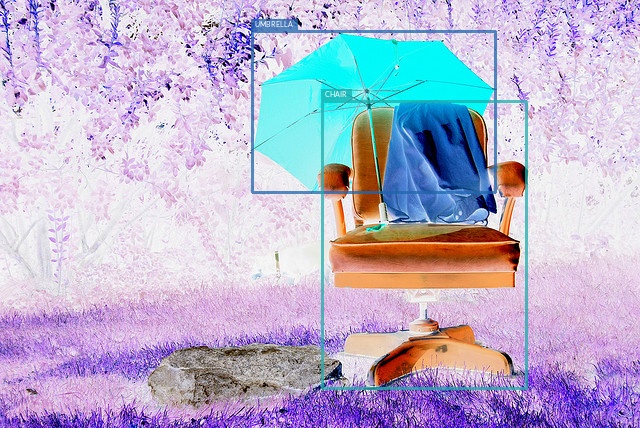}} & umbrella, chair & umbrella (0.99), chair (0.99), \textbf{dog (0.62), cat (0.41)} & umbrella(0.97), chair(0.97),   [\textbf{cat (0.01), dog (0.20)}] \\ \hline
\parbox[c]{1em}{\includegraphics[width=0.25\textwidth]{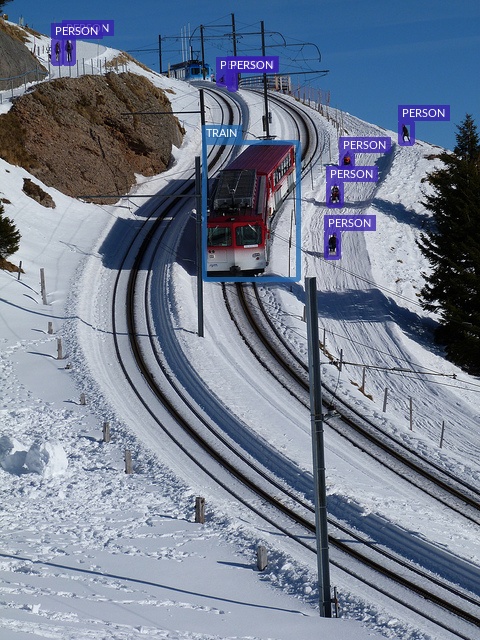}} & person, train & person (0.98), train (0.99), \textbf{bus (0.59), skis (0.55)} & person(0.99), train(0.97), [\textbf{bus   (0.11), skis (0.02)}] \\ \hline
\parbox[c]{1em}{\includegraphics[width=0.25\textwidth]{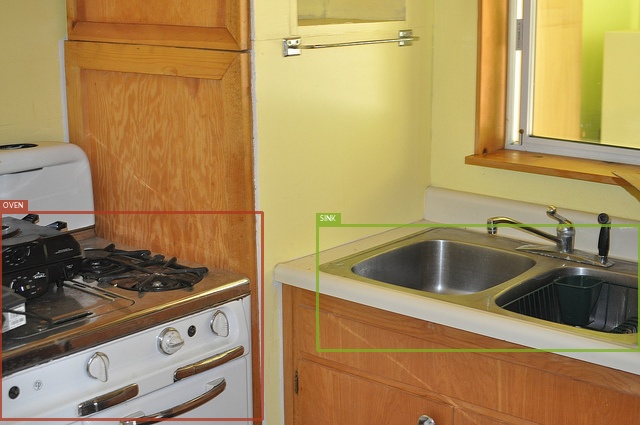}} & oven, sink & oven (0.99), sink (0.99),   \textbf{refrigerator (0.60), toaster (0.48)} & sink(0.97), oven(0.95), [\textbf{toaster   (0.02), refrigerator (0.05)}] \\ \hline
\parbox[c]{1em}{\includegraphics[width=0.25\textwidth]{images/ddn_vs_cnn/ddn_worse/COCO_val2014_000000002985.jpg}} & car, truck, elephant & car (0.99), truck (0.99), elephant (0.99),   [\textbf{person (0.22)}] & \textbf{person(0.42)}, car(0.98),   truck(0.97), elephant(0.97) \\ \hline
\parbox[c]{1em}{\includegraphics[width=0.25\textwidth]{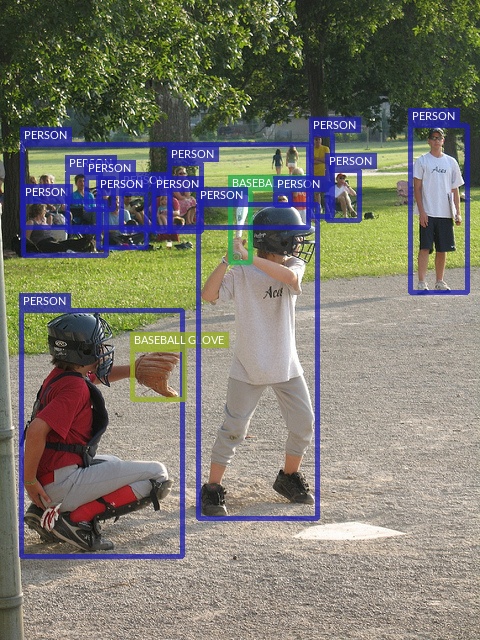}} & person, baseball bat, baseball   glove & person (0.99), baseball bat   (0.99), baseball glove (0.99), [\textbf{sports ball (0.25), chair (0.28)}] & person(0.99), \textbf{sports ball(0.47)},   baseball bat(0.97), baseball glove(0.97), \textbf{chair(0.51)} \\ \hline
\parbox[c]{1em}{\includegraphics[width=0.25\textwidth]{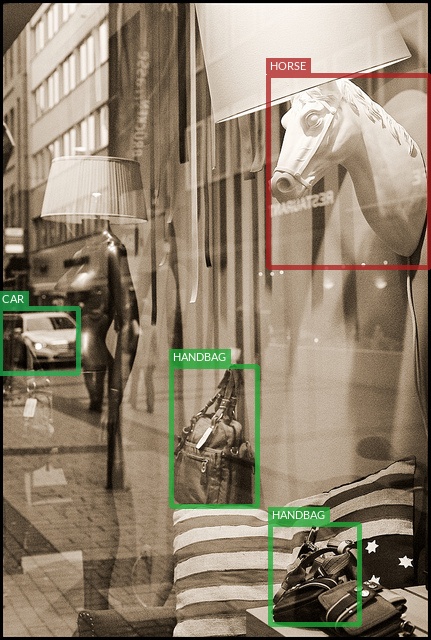}} & car, horse, handbag & car (0.99), horse (0.99),   handbag (0.99), [\textbf{person (0.20)}] & \textbf{person(0.58)}, car(0.98),   horse(0.97), handbag(0.97) \\ \hline
\parbox[c]{1em}{\includegraphics[width=0.25\textwidth]{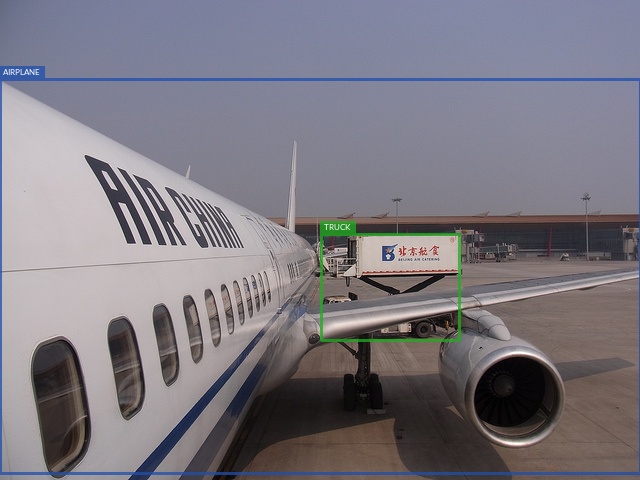}} & airplane, truck & airplane (0.99), truck (0.90),   [\textbf{person (0.20)}] & \textbf{person(0.34)}, airplane(0.98),   truck(0.86) \\ \hline
\parbox[c]{1em}{\includegraphics[width=0.25\textwidth]{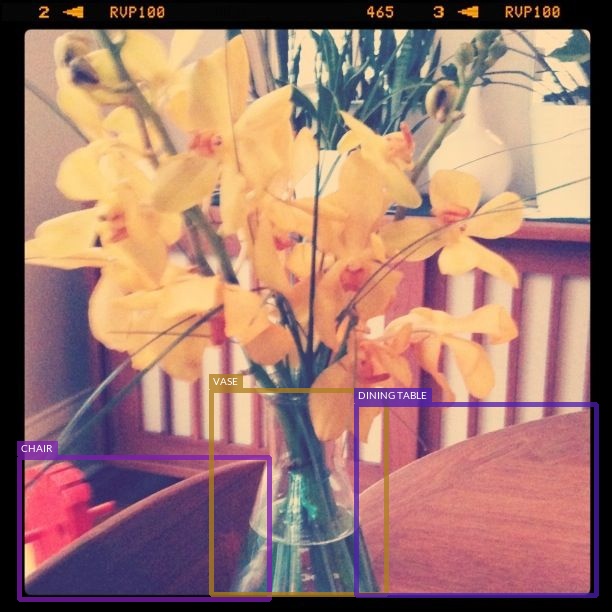}} & chair, dining table, vase & chair (0.93), dining table   (0.98), vase (0.99), [\textbf{potted plant (0.26)}] & chair(0.93), \textbf{potted plant(0.45)},   dining table(0.96), vase(0.97) \\ \hline
\parbox[c]{1em}{\includegraphics[width=0.25\textwidth]{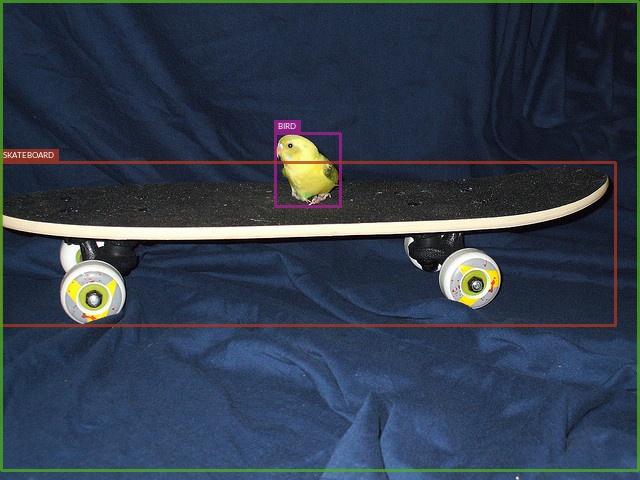}} & bird, skateboard, couch & bird (0.99), skateboard (0.99),  \textbf{couch (0.73)} & bird(0.98), skateboard(0.94),   [\textbf{couch (0.15)}] \\ \hline
\parbox[c]{1em}{\includegraphics[width=0.25\textwidth]{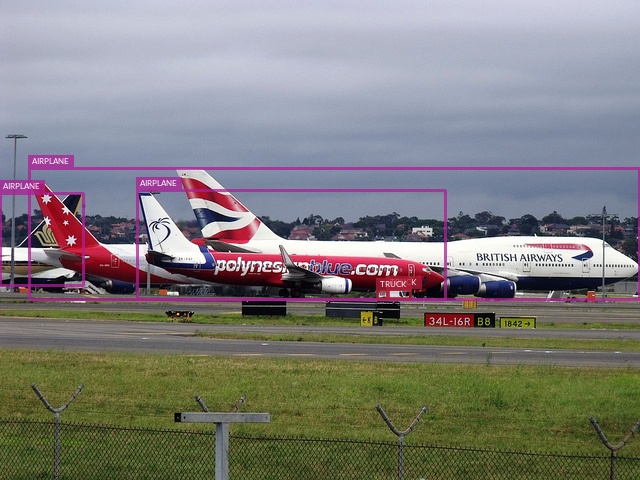}} & airplane, truck & airplane (0.99), \textbf{truck (0.42)} & airplane(0.98), [\textbf{truck (0.27)}] \\ \hline
\parbox[c]{1em}{\includegraphics[width=0.25\textwidth]{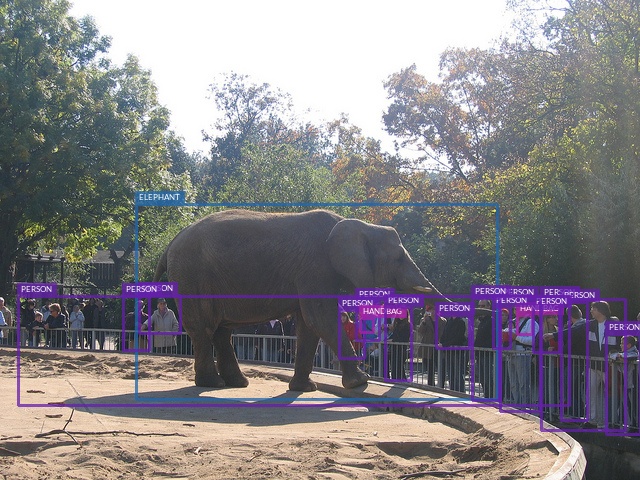}} & person, elephant, handbag & person (0.99), elephant (0.99),  \textbf{handbag (0.36)} & person(0.99), elephant(0.97),   [\textbf{handbag (0.23)}] \\ \hline
\parbox[c]{1em}{\includegraphics[width=0.25\textwidth]{images/ddn_vs_cnn/ddn_worse/COCO_val2014_000000021161.jpg}} & cat, sandwich, bed & cat (0.99), \textbf{sandwich (0.83)}, bed   (0.99) & cat(0.98), bed(0.97), [\textbf{sandwich   (0.27)}] \\ \hline
\parbox[c]{1em}{\includegraphics[width=0.25\textwidth]{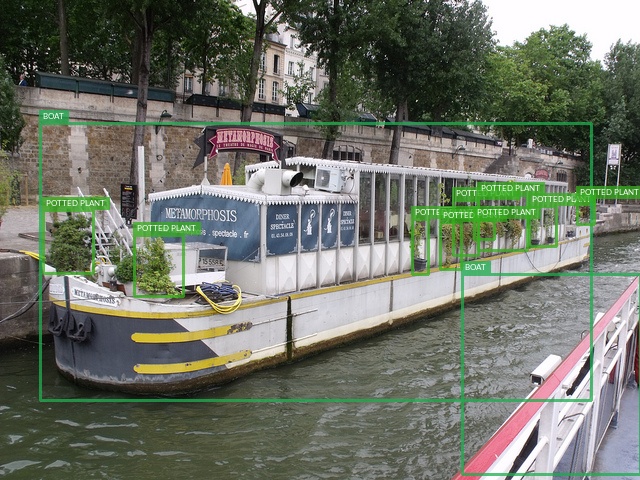}} & boat, potted plant & boat (0.99), \textbf{potted plant (0.50)} & boat(0.97), [\textbf{potted plant   (0.11)}] \\ \hline
\end{longtable}

\end{document}